\documentclass{article}

\PassOptionsToPackage{numbers,sort&compress}{natbib}
\usepackage[main, final]{neurips_2026}
\usepackage{wrapfig}
\usepackage{graphicx}
\usepackage{amsmath}
\usepackage{amsthm}
\usepackage{amssymb}
\usepackage{booktabs}
\usepackage{xcolor}
\usepackage{tikz}
\usetikzlibrary{calc,positioning}
\usepackage[colorlinks=true,urlcolor=blue]{hyperref}
\usepackage{fontawesome5}
\usepackage{subcaption}
\usepackage{placeins}

\usepackage[normalem]{ulem}
\usepackage{xcolor}

\newtheorem*{restatedproposition}{Proposition}
\newtheorem{proposition}{Proposition}

\newcommand{\ours}{KEAP}

\title{Know What You Need:\\Efficient Activation Checkpointing}

\author{
  Danyal Ahmed \\
  Sorbonne University\\
  Paris, France \\
  \texttt{danyal.ahmed@sorbonne-universite.fr}
  \And
  Edouard Oyallon \\
  CNRS, Sorbonne University\\
  Paris, France
}
\date{} 
\begin{document}

\maketitle

\begin{abstract}
Large Language Model training is increasingly constrained by memory capacity. Activation Checkpointing reduces activation memory by avoiding storing selected activations during the forward pass and rematerializing them during backpropagation, thereby trading memory savings for additional computation. Although this technique is essential for scaling model training, rematerialization can be prohibitively expensive, particularly for Transformer linear layers. Approximate computation could reduce this overhead, but conventional approaches either introduce errors that  accumulate over the backward pass or fail to exploit information available from the initial forward pass. We introduce a novel approximate rematerialization method that can yield unbiased gradient estimates:
 using information collected during the initial forward pass, the method selectively recomputes only the most important layer components, either through greedy selection or randomized sketching. Ablation studies show that the resulting gradients closely match the exact gradients. When applied to feed-forward  layers, attention layers, or both, our method can recompute only two-thirds of the layer components in a Llama-488M model, while increasing perplexity by just 1.7\%, 1.2\%, and 2.5\%, respectively.
\end{abstract}

\section{Introduction}
Large Language Models (LLMs) have progressed largely through scaling, resulting in models with a larger number of parameters that need to process more data in parallel \cite{grattafiori2024llama,epochai2026parametertrends}. At the same time, accelerator throughput has improved much faster than memory capacity \cite{delsozzo2026progress}. This trend has made memory overhead a major bottleneck, particularly during gradient computation: while model states represent a constant fraction of the memory which can be sharded \cite{rajbhandari2020zero,singh2025model,fournier2024wash,nabli2026acco}, the activation memory also scales with batch size and context length \cite{xu2026128k}. This makes activations a particularly challenging source of memory pressure during training, since they must be stored during training for computing gradients. Indeed, these gradients are typically computed using reverse-mode automatic differentiation \cite{baydin2018automatic,linnainmaa1976taylor} over a computation graph, where operations are represented as nodes of this graph. During the forward pass, intermediate activations are stored and kept in memory until the backward pass, where they are used to compute gradients and are progressively released. Thus, this activation storage can represent a substantial memory overhead and must therefore be carefully managed.

Several strategies have been proposed to reduce this overhead, typically by trading memory savings for additional data movement or computation. One approach is to offload tensors, such as model states \cite{ren2021zero} or activations \cite{rhu2016vdnn}, from the accelerator to CPU memory. However, this can be inefficient in practice because of the high cost of repeated data transfers between devices. Another approach is to carefully schedule pipeline execution to reduce peak memory usage~\cite{janson2026learned}, but such schedules often introduce their own drawbacks, such as  communication overhead, known as pipeline bubbles \cite{huang2019gpipe}, and load imbalance across layers \cite{kim2023bpipe}, since earlier pipeline stages typically retain activation memory for longer as they are the last to process gradients. A notable example is the V-shaped pipeline schedule used in \cite{liu2024deepseek}, which improves memory efficiency under specific settings but comes with additional scheduling constraints.

A natural way to avoid storing activations is to recompute them on the fly during the backward pass. For instance, instead of storing activations during the forward pass, invertible architectures \cite{jacobsen2018revnet,gomez2017reversible,rivaud2025petra} use model invertibility to progressively reconstruct them while simultaneously computing gradients during the backward pass, at the cost of additional computation. This idea of reconstructing activations is closely related to Activation Checkpointing (AC) \cite{griewank1992achieving}, which also recomputes selected parts of the computation graph during the backward pass.  Any missing activations required during backpropagation are recomputed on the fly from the latest available checkpoint, enabling gradient computation with lower memory usage at the cost of additional computation, which can cause up to a $33\%$  compute overhead  compared to a vanilla backward pass. Consequently, AC trades memory overhead for computation,  but its benefits are layer-dependent: selective AC \cite{korthikanti2023reducing} allows one to choose the layers for which this trade-off is favorable. For instance, thanks to its IO-aware implementation, FlashAttention \cite{dao2022flashattention} recomputes the attention module during the backward pass, with a negligible overhead: the most important intermediate quantities are rematerialized in the accelerator cache. In contrast, linear layers typically retain matrix-multiplication outputs, since these operations are expensive to rematerialize during backpropagation.

In this work, we challenge the need to fully recompute expensive linear layers to obtain the memory benefits of Full AC. In contrast to existing approaches that approximate both the forward and backward pass or the gradient computation alone \cite{micikevicius2018mixed, gray2017blocksparse, sun2017meprop}, we propose an approximation strategy tailored specifically to AC: we approximate the computations performed during rematerialization. Accordingly, the approximation we use stems from the observation that many monolithic computational blocks can be expressed as linear combinations of elementary components, and can therefore be effectively approximated through sketching \cite{woodruff2014sketching}. \textit{Consequently, only a fraction of the activations produced by linear layers need to be rematerialized  by AC.} Since checkpoints are stored exactly, the approximation is repeatedly reset to reliable intermediate states, preserving unbiased gradient estimates and preventing errors from propagating over long computation chains.  We propose \ours{} (\textbf{K}now What You Need: \textbf{E}fficient \textbf{A}ctivation Check\textbf{p}ointing), an approximation method that leverages  activation checkpointing by selectively approximating the computations needed during recomputation, thanks to an initial exact forward pass.

\paragraph{Contributions.} We make the following contributions. \textbf{(1)} We introduce a novel AC strategy, \ours, that approximates rematerialization using only information obtained from the full forward pass. \textbf{(2)} We show how to maintain high-quality gradients, either through unbiased estimation or through a greedy approximation scheme. \textbf{(3)} We explain how to apply the method to both FFN and attention layers. \textbf{(4)} We demonstrate numerically, on up to 488M parameter Llama-style models, that \ours{} retains the activation-memory benefits of full AC while causing only limited perplexity degradation. \textbf{(5)} We provide ablation studies to better understand the trade-off between AC approximation quality and model accuracy.

\paragraph{Structure of the paper.} Sec.~\ref{sec:main-method} introduces the necessary background on AC, while Sec.~\ref{sec:main-metod-kweak} shows how \ours{} can recover unbiased gradients, followed by a discussion on the computational benefits in Sec.~\ref{sec:implem}. Sec.~\ref{sec:num-exp} presents our numerical experiments, followed by ablation studies in Sec.~\ref{sec:ablations}. Our source code can be found at \href{https://github.com/da462/KEAP}{https://github.com/da462/KEAP}.

\section{Related Work}

\paragraph{Activation rematerialization.}
A natural way to avoid storing activations is to recompute them during the backward pass \cite{griewank1992achieving}. Several works study optimal or near-optimal rematerialization strategies, for instance through checkpointing schedules \cite{griewank2000algorithm,beaumont2024optimal,gusak2025hiremate,aguila2025optimized}. Closely related to this line of work, \citet{chen2025adacc} improves AC by adaptively deciding, for each activation tensor, whether it should be recomputed during the backward pass or stored in compressed form, thereby reducing memory usage while limiting recomputation overhead. Invertible neural networks provide another way to reduce activation storage by reconstructing activations during the backward pass, in particular through lifting-based reversible architectures \cite{gomez2017reversible,jacobsen2018revnet}. Similar to standard AC, such reversible blocks can also be combined with checkpointing when only part of the network is reversible \citep{rivaud2025petra}.

\paragraph{Approximating activations for backward computation.}
Combinable with AC, approximate-activation methods store compressed or approximate versions of activation states during the forward pass to save memory. \citet{chakrabarti2019backprop} showed that using compressed activations has limited impact on training quality. \citet{liu2022gact} extend this idea to the AC setting, where rematerialization starts from quantized checkpoints, without trying to reduce computational footprint. In contrast to \ours{}, the local Jacobians are then evaluated at approximate rather than exact forward activations, allowing approximation errors to propagate through the backward computation.  Relatedly, \citet{kong2025clapping} reduce per-sample storage in pipeline-parallel training through communication compression, while \citet{shamshoum2025compact} compress activations during the forward pass and reuse them during backpropagation. However, these methods typically decode compressed activations during the backward pass and therefore mainly reduce memory, not recomputation. In contrast, \ours{} performs only a subset of the original rematerialization computations, for both parameter-gradient and input-gradient computations during the backward pass.

\paragraph{Approximate backpropagation computation.}
Approximate-gradient methods reduce the cost of the backward pass directly, and could also be combined with AC. \citet{sun2017meprop} exploit the empirical sparsity of per-example gradients by keeping only the top-\(k\) entries of the incoming gradient, yielding a biased lower-FLOP backward pass. \citet{chmiel2023mvue} instead enforce unbiasedness for sparse backpropagated gradients and show that the random pruning mask should minimize variance rather than sparsification error~\citep{sarfi2025communication}. By partitioning the computational graph into independent segments, \citet{belilovsky2020decoupled} reduce activation-memory overhead, but at the cost of a  degraded accuracy. The local vector-Jacobian product itself has also been approximated with randomized unbiased estimators \citep{bakong2026unbiased}, although this still requires ad hoc computations and also degrades accuracy. In contrast, \ours{} approximates the rematerialization introduced by AC: the backward pass is computed through an approximate recomputation path rather than by directly modifying the error signal or the backward rule, while aiming to maintain a high quality gradient approximation.

\paragraph{Sketching strategies for neural networks.}
As we do, sketching \cite{woodruff2014sketching} has also been used to approximate neural-network computations. Useful to our work, \citet{daniely2017short} propose a theoretical framework for understanding sketches of one-hidden-layer neural networks, while \citet{gazagnadou2022ridgesketch} study related sketching ideas for ridge regression (a zero-hidden layer neural network). These works demonstrate the existence of useful sketches for forward computations. In a similar spirit, \citet{yang2022sketch} combine sketching with natural-gradient methods to reduce the cost of second-order optimization. Our setting differs in that sketching is applied specifically to the rematerialization path induced by AC. We use statistics from the first exact forward pass to remove components during  re-materialization. Due to their depth, deep neural networks are generally difficult to combine with sketching, as approximation errors would accumulate and propagate across many layers. Our method avoids this issue by applying sketching only within the recomputation path between exact checkpoints, leading to faithful approximations.

\section{Method}
\label{sec:main-method}
\subsection{Background on AC}
In the following, we represent the forward pass of a neural network (typically a transformer) as a cascade of nonlinear operators $\phi_j$ (that could be either a block of layers or a single layer), given by
\[
x_{j+1} = \phi_j(x_j,\vartheta_j),  \tag{Forward $j$}
\]
where \(\vartheta_j\) denotes the set of parameters of layer \(j\), with final loss $\mathcal{L}$. In standard backpropagation, the forward pass is followed, for each layer \(j\),  by the adjoint-state backpropagation
\[
\delta_j
= \left(J_{x_j} \phi_j(x_j,\vartheta_j)\right)^{\!\top}\delta_{j+1}
\tag{Backward $j$}
, \qquad
\frac{\partial \mathcal{L}}{\partial \vartheta_j}
= \left(J_{\vartheta_j} \phi_j(x_j,\vartheta_j)\right)^{\!\top}\delta_{j+1}\,.
\]
where $\delta_j = \frac{\partial \mathcal{L}}{\partial x_j}$ and $J_xf$ denotes the Jacobian of $f$ with respect to variable $x$. Sometimes, for the sake of simplicity, we may write $\phi_j(x_j)$   instead of $\phi_j(x_j,\vartheta_j)$.

\paragraph{AC.} Instead of storing all intermediate activations produced during the forward pass, only a selected subset is kept in memory, which we denote by $x^{\mathrm{AC}}_{j_1}, \ldots, x^{\mathrm{AC}}_{j_l}, \ldots $. The missing activations are then recomputed on demand during backpropagation. In this way, activation checkpointing trades additional computation for reduced memory usage. More precisely, let \(x^{\mathrm{AC}}_{j_l}\) be a stored checkpoint and introduce $\Phi_l=\phi_{j_{l+1}-1}\circ...\circ\phi_{j_l}$. Then, for any layer index \(j\) such that \(j_l < j \leq j_{l+1}\), the activation \(x_j\) is recovered by applying the corresponding forward operators from the checkpoint \(x^{\mathrm{AC}}_{j_l}\) via a second forward pass to recompute $\Phi_{l}(x^{\mathrm{AC}}_{j_l},\theta_l)$ with $\theta_l=(\vartheta_{j_l},..., \vartheta_{j_{l+1}-1})$, namely 
\begin{equation}
x^\mathrm{AC}_{j+1} = \phi_j(x^\mathrm{AC}_j,\vartheta_j),
\qquad j = j_l, \ldots, j_{l+1}-1. \tag{Recompute forward}
\label{eq:recomputeforward}
\end{equation}
Once the missing activations have been recomputed, the backward pass is applied as usual and those activations are progressively discarded, for $l = L, \ldots, 1$,

\[
\delta_{l}
=
\left(J_{x_{j_l}} \Phi_l(x^\mathrm{AC}_{j_l},\theta_l)\right)^{\!\top}\delta_{l+1},
\qquad
\frac{\partial \mathcal{L}}{\partial \theta_l}
=
\left(J_{\theta_l} \Phi_l(x^\mathrm{AC}_{j_l},\theta_l)\right)^{\!\top}\delta_{l+1}.\tag{Recompute backward}
\]

The indices $j_1,\ldots,j_l,\ldots$ are typically selected via
\emph{selective AC}, whereby only a subset of layers
retains additional intermediate quantities according to computational or memory
considerations, most commonly those produced by linear layers. \ours{}
challenges this prevailing practice.

\subsection{Toward \ours}\label{sec:main-metod-kweak}
We now introduce a modification of AC for separable layers. We consider layer blocks $\Phi=\Phi_l$  (potentially, a single layer), that decompose, for token sequences \(x=(x_t)_t\) so that $x_t\in\mathbb{R}^D$, as
\begin{equation}
\label{eq:sep-dec}
\Phi(x,\theta) = \sum_{i=1}^n \varphi^i(x,\theta^i).\tag{Sep}
\end{equation}
For instance,  feed-forward layers (FFN)   with  hidden dimension  $n$ expand componentwise as
\begin{equation}
\label{eq:FFN}
\mathrm{FFN}(x)_t
=
\sum_{i=1}^n
W^2_{:\!,i}\,
\Bigl(\mathrm{SiLU}((W^1x_t)_i)\,(W^3x_t)_i\Bigr),
\end{equation}

and likewise, a multi-head attention layer~\cite{vaswani2017attention} with $n$ heads can be written as a sum over heads
\begin{equation}
\label{eq:attention}
\mathrm{Attn}(x)_t=\sum_{i=1}^nW_i^o\,\mathbf{Att}(W_i^q x_t,\,W^k_i x,\,W^v_i x)\,.
\end{equation}

Note that any layer satisfies this property when $n=1$. Motivated by this separability property, we introduce sketching methods based on a masking vector, which may be either deterministic or stochastic. We define the corresponding sketched operator as

\begin{equation}
\label{eq:app-sep-dec}
\Phi^\omega(x,\theta) = \sum_{i=1}^n \omega^i \,\varphi^i(x,\theta^i) ,\tag{Approx Sep}
\end{equation}
where $\omega = (\omega^1,\dots,\omega^n)$ is a potentially $k$-sparse vector, meaning that it has only $k$ non-zero entries in expectation. A simple choice is to sample independently $z_i \sim \mathcal{B}(k/n)$ and set $\omega^i = \frac{n}{k} z_i$, so that $\mathbb{E}[\omega^i] = 1$ for $i=1,\dots,n$. In this case, the estimator is unbiased, since $\mathbb{E}[\Phi^\omega] = \Phi$. Its Jacobian estimator is also unbiased, as $\mathbb{E}[J\Phi^\omega] = J\Phi$ and, in expectation, only $k$ elements are computed. However, unbiasedness is generally not preserved under composition. Indeed, $$\mathbb{E}[J(\Phi_2^{\omega_2} \circ \Phi_1^{\omega_1})] \neq J(\Phi_2 \circ \Phi_1),$$ due to the dependence introduced by the forward pass, as highlighted by the blue term in
$$\mathbb{E}[J\Phi_2^{\omega_2}({\color{blue}{\Phi_1^{\omega_1}(x)}},\theta_2) \circ J\Phi_1^{\omega_1}(x,\theta_1)] \neq J\Phi_2(\Phi_1(x),\theta_2) \circ J\Phi_1(x,\theta_1)\,.$$

This issue is resolved by combining this approximation with AC. Indeed,  the first forward pass has already  constructed $x_2^{\mathrm{AC}} = \Phi_1(x_1^{\mathrm{AC}})$ with $x_1^{\mathrm{AC}}=x$, so that the Jacobians $J\Phi_1^{\omega_1}, J\Phi_2^{\omega_2}$ are recomputed via a second forward pass, so that
$$\mathbb{E}[J\Phi_2^{\omega_2}({\color{blue}{x_2^{\mathrm{AC}}}},\theta_2) \circ J\Phi_1^{\omega_1}(x_1^{\mathrm{AC}},\theta_1)] = J\Phi_2(\Phi_1(x),\theta_2) \circ J\Phi_1(x,\theta_1)\,.$$
This leads to the following generalization, whose proof can be found in App.~\ref{app:proof}:
\begin{proposition}[Unbiased gradient in the separable setting]
\label{prop:unbiased-ac-ours}
Let $
    \mathcal{L}\circ \Phi_L \circ \cdots \circ \Phi_1
$ be a model and assume that each block $\Phi_l$ admits a separable decomposition as in
Eq.~\eqref{eq:sep-dec}. For each block, let $\omega_l$ be a random mask such
that $\mathbb{E}[\omega_l]=\mathbf{1}$, with masks sampled independently across
blocks. Let $\{x_l^{\mathrm{AC}}\}_{l\leq L}$ be the checkpointed activations obtained from the exact forward pass, as in
Eq.~\eqref{eq:recomputeforward}. Then, the sketched AC backward pass gives an
unbiased  estimator
\[
\mathbb{E}_{\omega}
\left[
\left(
\widehat{\nabla}_{\theta_1}^{\,\omega,\mathrm{AC}}\mathcal{L},
\ldots,
\widehat{\nabla}_{\theta_L}^{\,\omega,\mathrm{AC}}\mathcal{L}
\right)
\right]
=
\nabla_{(\theta_1,\ldots,\theta_L)}
\mathcal{L}\,,
\]
where we have for $1\leq l\leq L$,
\[
\widehat{\nabla}_{\theta_l}^{\,\omega,\mathrm{AC}}\mathcal{L}
:=
\left(
J_{\theta_l}\Phi_l^{\omega_l}(x_l^{\mathrm{AC}},\theta_l)
\right)^{\!\top}
\left(
J_x\Phi_{l+1}^{\omega_{l+1}}(x_{l+1}^{\mathrm{AC}},\theta_{l+1})
\right)^{\!\top}
\cdots
\left(
J_x\Phi_L^{\omega_L}(x_L^{\mathrm{AC}},\theta_L)
\right)^{\!\top}
\nabla_{x_{L+1}}\mathcal{L}.
\]
\end{proposition}

Building on this observation, we introduce three strategies within the \ours{} framework:

\paragraph{Rand-$\frac{k}n$.}
A simple and naïve baseline is to sample $z \in \{0,1\}^n$ uniformly with exactly $k$ ones, and set
$\omega^i = \frac{n}{k} z_i$. In this case, we have $\mathbb{E}[J\Phi^\omega]=J\Phi$.
\paragraph{Opt-$\frac{k}n$.}We optimize the sampling probabilities $\{p_i\}_{i\leq n}$ under an expected budget of $k$ selected components. Specifically, we sample independent Bernoulli variables $z_i \sim \mathcal{B}(p_i)$ with $\omega^i=\frac {z_i}{p_i}$, while enforcing a lower bound $\varepsilon$ to avoid numerical instabilities. For samples $\{x^m\}_m$, it leads to
\begin{equation}
\label{eq:optimal-sampling-problem}\tag{*}
\begin{aligned}
    \min_{\{p_i\}_{i=1}^n}\quad
    & \sum_m \mathbb{E}_\omega
    \bigl\|\Phi(x^m)-\Phi^\omega(x^m)\bigr\|^2 \\
    \mathrm{s.t.}\quad
    & \sum_{i=1}^n p_i = k,
    \qquad
    \varepsilon \le p_i \le 1,\quad i=1,\ldots,n .
\end{aligned}
\end{equation}
\paragraph{Top-$\frac kn$.}
For a collection of inputs $\{x^m\}_m$, define $\varphi^i:=\{\varphi^i(x^m,\theta^i)\}_{ m}$, so that  $\|\varphi^i\|^2
    =
    \sum_m\|\varphi^i(x^m,\theta^i)\|^2.$ In this deterministic strategy, we keep only the $k$ components with largest norm $\|\varphi^i\|$, and set the remaining ones to zero. Equivalently, $\omega^i=1$ for the $k$ indices corresponding to the largest values of $\|\varphi^i\|$, and $\omega^i=0$ otherwise. This yields a biased but better approximation by preserving only the dominant contributions to $\Phi(x,\theta)$.

The following proposition solves the Opt-$\frac{k}{n}$ problem in Eq.~\eqref{eq:optimal-sampling-problem}; the proof is deferred to App.~\ref{app:proof}.

\begin{proposition}[Optimal Bernoulli sampling probabilities]
\label{prop:opt-bernoulli}
Let $\omega^i=\frac{z_i}{p_i}$ and $z_i\sim \mathrm{Bernoulli}(p_i)$ independent, and let \(\Phi\) and \(\Phi^\omega\) be defined as in
Eq.~\eqref{eq:sep-dec} and Eq.~\eqref{eq:app-sep-dec}, respectively. Then the probabilities $p_i$ minimizing Eq. \eqref{eq:optimal-sampling-problem} are given by
\[
    p_i^\star
    =
    \min\!\left(
        1,
        \max\!\left(
            \varepsilon,
            \frac{\|\varphi^i\|}{\lambda}
        \right)
    \right),
\]
where $\lambda>0$ is chosen so that
$
    \sum_{i=1}^n
    \min\!\left(
        1,
        \max\!\left(
            \varepsilon,
            \frac{\|\varphi^i\|}{\lambda}
        \right)
    \right)
    = k$.
In particular, if the box constraints are inactive, i.e., $
    \varepsilon
    \le
    \frac{k\|\varphi^i\|}
         {\sum_{j=1}^n \|\varphi^j\|}
    \le 1$
    for all  $i$,
 then
$
    p_i^\star
    =
    \frac{k\|\varphi^i\|}
         {\sum_{j=1}^n \|\varphi^j\|}.$
\end{proposition}In fact, we can also demonstrate a pseudo-optimality
of Top-$\frac kn$. For a mask $\omega$, let
\[
Q(\omega)\triangleq\mathbb{E}_\omega\sum_m\bigl\|\Phi(x^m)-\Phi^\omega(x^m)\bigr\|^2 ,
\]
and introduce the
coherence of the discarded components
\[
\mu
\triangleq
\max_{\substack{i\neq j\\ i,j\notin \mathrm{Top}_k}}
\left[
\frac{\langle \varphi^i,\varphi^j\rangle}{\|\varphi^i\|\,\|\varphi^j\|}
\right]_+ .
\]
The following proposition shows that an arbitrarily small error for Opt-$\frac{k}{n}$ implies a correspondingly small error for Top-$\frac{k}{n}$, whereas the converse does not hold, thereby highlighting the advantage of Top-$\frac{k}{n}$.This proposition bounds the approximation error of a single block but does not guarantee unbiased gradients. Nevertheless, our ablations show that Top-$\frac{k}{n}$ maintains high gradient similarity in practice (Fig.~\ref{fig:main-ablation_cosine}).

\begin{proposition}[Top-$k$ versus optimal Bernoulli sampling]
\label{prop:top-vs-opt}
Let $\omega^{\mathrm{Top}}$ and $\omega^{\mathrm{Opt}}$ be the masks of Top-$\frac kn$ and
Opt-$\frac kn$. Then
\[
Q(\omega^{\mathrm{Top}})
\;\le\;
\bigl(1+(n-k-1)\mu\bigr)\!\!\sum_{i\notin \mathrm{Top}_k}\!\!\|\varphi^i\|^2
\;\le\;
\bigl(1+(n-k-1)\mu\bigr)\,Q(\omega^{\mathrm{Opt}}).
\]
In particular $Q(\omega^{\mathrm{Top}})\le(n-k)\,Q(\omega^{\mathrm{Opt}})$ since $\mu\le1$,
and $Q(\omega^{\mathrm{Top}})\le Q(\omega^{\mathrm{Opt}})$ whenever $\mu=0$. Conversely, no reverse bound holds:
for $n-k\ge2$ there is no constant $C(n,k)$ such that
$Q(\omega^{\mathrm{Opt}})\le C(n,k)\,Q(\omega^{\mathrm{Top}})$.
\end{proposition}

\subsection{Implementation of \ours}\label{sec:implem}

\begin{figure}[t]
\centering
\includegraphics[width=\linewidth]{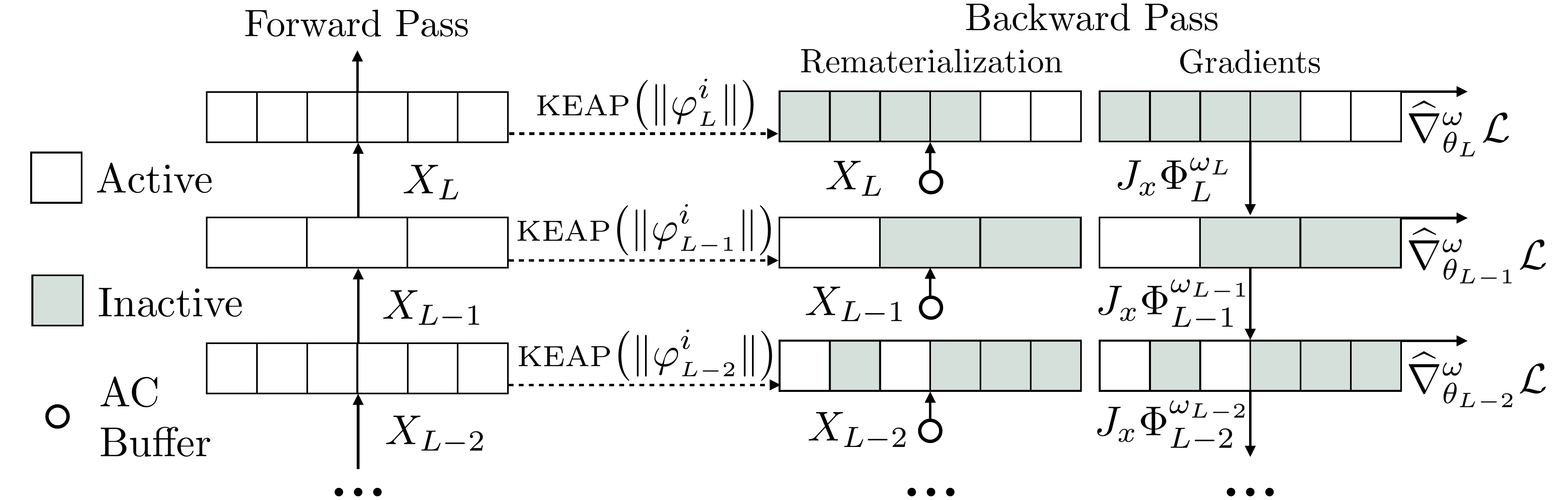}
\caption{\textbf{High-level overview of \ours.} During an initial forward pass, all components $\varphi^i$ of a layer are computed. \ours\, then selects the most important components according to their $\ell^2$ norms. During rematerialization, only this selected fraction  is recomputed and used for the backward pass.}
\label{fig:KWEAK-diagram}
\end{figure}

We note that both Opt-$\frac kn$ and Top-$\frac kn$ require the statistic $\|\varphi^i\|$, which can be computed during the first forward pass at negligible additional cost, and even overlapped with other computations, which is an advantage of our method. The \ours{} strategy is illustrated in Fig.\,\ref{fig:KWEAK-diagram}. We next compare the FLOP and activation-memory costs of \ours{} with vanilla backpropagation and standard activation-checkpointing strategies. These FLOP reductions provide a useful reference for computational savings. To better realize these gains in wall-clock time, we also introduce strategies that group our approximation across tokens and channels, enabling larger matrix multiplications; we evaluate these strategies separately in App.~\ref{sec:main-block}.

\paragraph{Vanilla Backpropagation.} We analyze the vanilla backward pass on a per-token basis using
Eq.~\eqref{eq:FFN} and Eq.~\eqref{eq:attention}. For an FFN with hidden
dimension $n$, the forward and backward costs are $3 \times 2nD$ and
$3 \times 4nD$ FLOPs, respectively. In Llama-style models, one typically
has $n=\frac{8}{3}D$~\citep{grattafiori2024llama}. For multi-head attention with $n$ heads of dimension $d$, each head
projects tokens to a $d$-dimensional subspace, typically with $D=nd$.
Per head and token, the forward and backward costs are $4 \times 2dD$
and $4 \times 4dD$ FLOPs, respectively. We neglect the cost of the
attention operation itself compared to matrix multiplications, as it is
limited in practice~\citep{dao2022flashattention}. The activation memory stored during the forward pass depends on the backpropagation implementation, which is sometimes combined with AC-like strategies without being stated explicitly~\cite{paszke2019pytorch}. 

\paragraph{Vanilla AC.}We now describe AC applied to a full FFN or attention layer, as defined in Eq.~\eqref{eq:FFN} or Eq.~\eqref{eq:attention}.
For the FFN, AC introduces a computational overhead of $3 \times 2nD$, while reducing the memory footprint to solely $D$. For the attention layer, the overhead is $4 \times 2ndD$  and the memory footprint is also reduced to $D$. 

\paragraph{Selective AC.}
Selective AC generally keeps all activations resulting from linear computations \cite{korthikanti2023reducing,pytorch_activation_checkpointing_2025}. Thus, for FFN and attention layers, the memory footprint per token becomes $2n+2D$ and $5D$, respectively, but no additional recomputation overhead is introduced. As above, we neglect the memory of the attention operation itself. This strategy is typically preferred over vanilla AC to reduce computational overhead. 

\paragraph{\ours.} The goal of \ours\,is to combine the memory benefit of Vanilla AC while recomputing only a fraction of the layer components. For  FFN, the activation memory footprint is reduced to $D$, as for Vanilla AC, and the total cost per token becomes $3\times2nD +3\times 2kD + 3 \times 4kD=6nD+18\frac knnD$. For attention, if we skip entire heads, the memory footprint is reduced to $D$ and the total cost per token becomes $4\times 2ndD+4\times 2kdD+4\times 4kdD=8D^2+24\frac knD^2$. This is done, for instance, using batch-level approximation granularity.
However, this may increase variance of our gradient approximation. If masks are estimated at token-level
granularity, the full keys and values must be stored, as shown in Eq.~\eqref{eq:attention}.
Although they could be recomputed on the fly like the other activations, we
instead store them to avoid additional computational overhead. Consequently, the total cost per token becomes $4\times 2ndD+2\times 2kdD+2\times 4kdD+2\times 4ndD=16D^2+12\frac knD^2$ but the memory footprint per token increases to $D+2nd=3D$, still smaller than $5D$.

\section{Numerical Experiments}
\label{sec:num-exp}
\subsection{Training setup}
\label{sec:training-setup}

\paragraph{Architecture hyperparameters.}
We closely follow the tuned baselines described in
\citep{janson2026stabilizing}. All models use a
Llama-3-style architecture with RoPE positional embeddings
\citep{su2024roformer}. We use the Llama-2 7B
tokenizer, with vocabulary size \(32\mathrm{k}\) because it incurs a
smaller memory overhead compared to the Llama-3 tokenizer. We did not observe any significant discrepancy
relative to the Llama-3 tokenizer family. We consider two model configurations. The first is a 134M-parameter model
with hidden size 768, 12  layers, and 12 attention heads, which is used
for ablation studies. The second is a 488M-parameter model with hidden size
1280, 20 layers, and 20 attention heads, which is used for the main
numerical results in Sec.~\ref{sec:main-results}. Indeed, this model scale provides a practical compromise between computational cost and empirical relevance, making it suitable for assessing the potential scalability of our method \cite{hoffmann2022training}.

\begin{table*}[t]
\centering
\small
\setlength{\tabcolsep}{3pt}
\centering
\begin{subtable}[t]{0.45\textwidth}
\begin{tabular}{lcccc}
\toprule
Method &  $\frac kn$ & AC-FLOPs & PPL & $\Delta$PPL \\
\midrule
\multicolumn{5}{l}{\footnotesize AC applied on FFN layers} \\
\midrule
Llama-488M &  1    & 1.33 & 12.03 & -- \\
Llama-419M   & 1    & 1 & 12.44 & 3.41\% \\
\midrule
Rand    &   0.66 & 1  & 12.84  & 6.73\%\\
        &   0.33 & 0.66 & 14.70 & 22.19\% \\
\midrule
Opt     &   0.66 & 1    & 12.58 & 4.57\% \\
        &   0.33 & 0.66 & 13.66 & 13.55\% \\
\midrule
Top     &   0.66 & 1 & \textbf{12.23} & \textbf{1.66\%}\\
        &   0.33 & 0.66 & 12.46 & 3.57\%  \\
\midrule
\multicolumn{5}{l}{\footnotesize AC applied on Attention layers} \\
\midrule
Llama-488M &   1    & 1.33 & 12.03 & -- \\
Llama-468M   & 1    & 1  & \textbf{12.17} & \textbf{1.16\%} \\
\midrule
Rand    &   0.66 & 1  & 12.55 & 4.32\% \\
        &   0.33 & 0.83 & 13.43 & 11.64\%   \\
\midrule
Opt     &   0.66 & 1 & 12.27 & 2.00\%\\
        &   0.33 & 0.83  & 12.74 & 5.90\% \\
\midrule
Top     &   0.66 & 1 & \textbf{12.17} & \textbf{1.16\%}\\
        &   0.33 & 0.83 & 12.45 & 3.49\% \\
\midrule
\multicolumn{5}{l}{\footnotesize AC applied on Attention+FFN layers} \\
\midrule
Llama-488M &   1    & 1.33 & 12.03 & -- \\
Llama-400M   & 1    & 1  & 12.56 & 4.41\%  \\
\midrule
Rand    &   0.66 & 1 & 13.47 & 11.97\% \\
        &   0.33 &  0.72 & 17.91 & 48.88\%  \\
\midrule
Opt     &   0.66 & 1 & 12.81 & 6.48\% \\
        &   0.33 & 0.72 & 14.82 & 23.19\% \\
\midrule
Top     &   0.66 & 1 & \textbf{12.33}& \textbf{2.49\%} \\
        &   0.33 & 0.72 & 12.90 & 7.23\%\\
\bottomrule
\end{tabular}
\caption{Token-level granularity}
\label{tab:token-level}
\end{subtable}
\hfill
\begin{subtable}[t]{0.45\textwidth}
\begin{tabular}{lcccc}
\toprule
Method & $\frac kn$ & AC-FLOPs & PPL & $\Delta$PPL \\
\midrule
\multicolumn{5}{l}{\footnotesize AC applied on FFN layers} \\
\midrule
Llama-488M  & 1    & 1.33 & 12.03 & -- \\
Llama-419M  &  1    & 1 & \textbf{12.44} & \textbf{3.41\%} \\
\midrule
Rand    &   0.66 & 1 & 13.06 &  8.56\%  \\
        &   0.33 & 0.66 & 15.55 & 29.26\% \\
\midrule
Opt     &   0.66 & 1 & 12.89 & 7.15\%  \\
        &   0.33 & 0.66 & 14.30 & 18.87\%\\
\midrule
Top     &   0.66 & 1 & 12.61 & 4.82\%  \\
        &   0.33 & 0.66 & 13.40 & 11.39\% \\
\midrule
\multicolumn{5}{l}{\footnotesize AC applied on Attention layers} \\
\midrule
Llama-488M &   1    & 1.33 & 12.03 & -- \\
Llama-456M   & 1    & 1 & \textbf{12.22} & \textbf{1.58\%} \\
\midrule
Rand    &   0.66 & 1 & 12.97 & 7.81\% \\
        &   0.33 & 0.66 & 14.47 & 20.28\%\\
\midrule
Opt     &   0.66 & 1 & 12.79 & 6.32\%\\
        &   0.33 & 0.66 & 13.77 & 14.46\% \\
\midrule
Top     &   0.66 & 1 & 12.35 & 2.66\%  \\
        &   0.33 & 0.66 & 12.54 & 4.24\% \\
\midrule
\multicolumn{5}{l}{\footnotesize AC applied on Attention+FFN layers} \\
\midrule
Llama-488M &   1    & 1.33 & 12.03 & -- \\
Llama-387M   & 1    & 1 & \textbf{12.59} & \textbf{4.66\%} \\
\midrule
Rand    &   0.66 & 1 & 14.25 & 18.45\% \\
        &   0.33 & 0.66 & 20.23 & 68.16\% \\
\midrule
Opt     &   0.66 & 1    & 13.80 & 14.71\% \\
        &   0.33 & 0.66 & 17.06 & 41.81\% \\
\midrule
Top     &   0.66 & 1    & 12.85 & 6.82\% \\
        &   0.33 & 0.66 & 14.14 & 17.54\% \\
\bottomrule
\end{tabular}
\caption{Batch-level granularity}
\label{tab:batch-level}
\end{subtable}

\caption{\textbf{\ours{} performance}.
We report the perplexity of a Llama-488M model under different AC strategies, using per-token (\textbf{a}) and per-batch (\textbf{b}) granularity  during re-computation as described in Sec.~\ref{sec:implem}. $\Delta$ PPL is the relative perplexity degradation compared to the vanilla baseline. AC-FLOPs denotes the analytic FLOP cost of the targeted FFN and attention-projection computations relative to standard backpropagation for the same modules. (details can be found in  App.~\ref{sec:app-flops}).
}

\label{tab:remat-results}
\end{table*}

\paragraph{Training hyperparameters.}
We follow standard training practice and tune the baselines using a
linear-warmup cosine-decay learning-rate schedule
\citep{loshchilov2017sgdr}. Warmup is applied
during the first 5\% of training steps, followed by cosine decay over the
remaining steps, with the learning rate decaying to zero. The tuning sweep
covers learning rates over three logarithmic scales,
$\{10^{-2}, 10^{-3}, 10^{-4}\}$, with linear multipliers $\{1,5,7\}$, and
weight-decay values ranging from $10^{-1}$ to $10^{-3}$
\citep{loshchilov2019decoupled}. This sweep allows
us to cross-validate the best learning rate and weight decay. We use the Muon optimizer \citep{keller2024muon}, which we found to provide
good stability, with its standard settings: momentum \(0.95\) with Nesterov
momentum. We also  clip
the gradient norm to $1.0$. All models are trained on FineWeb
\citep{penedo2024fineweb}, which contains approximately  a total of 15T tokens.
We use batch size \(512\) and sequence length \(2048\), corresponding to
\(1{,}048{,}576\) tokens per training step. Dense baselines are trained to
Chinchilla-optimal token counts \citep{hoffmann2022training}, corresponding
to 2555 and 9307 training steps for the 134M and 488M models, or 2.68B and 9.76B processed tokens, respectively. In practice, the number of attention heads is small ($n\sim 10^2$) compared
to the FFN hidden dimension ($n\sim 10^3$), so we apply the clipping strategy
of Prop.~\ref{prop:opt-bernoulli} only to FFN layers. Numerically, setting
$\epsilon=0.06$ gives good performance without loss explosion.

\paragraph{Evaluation protocol.}
We report corresponding  perplexity (PPL) on a held-out FineWeb evaluation set containing
25M tokens (disjoint from the training set), and we also report the evaluation loss in App.~\ref{app:eval}. To evaluate the impact of our procedure, we compare it against two baselines: a reference model and a compute-matched  AC baseline.  AC does not change the model’s backward computation, but it increases the FLOPs of rematerialized modules by 33.3\%. Therefore, the compute-matched AC baseline uses a smaller model whose total training FLOPs match the  budget considered in App.~\ref{sec:app-flops}. Specifically, when AC is applied only to FFN layers, we reduce the FFN hidden dimension in the baseline. When AC is applied only to attention layers, we reduce the number of attention heads $n$. When \ours{} is applied to both FFN and attention layers, we apply both reductions while keeping the model width $D$ fixed. An alternative would be to apply a global model width multiplier~\cite{tay2022scale,yang2021tuning,hoffmann2022training}, but we did not observe significant differences with this approach. This reduction leads to slight FLOP differences for attention layers depending on whether token- or batch-level granularity is used; we discuss this further in App.~\ref{sec:app-flops}. We also tune the learning rate for the compute-matched AC baselines and find that the selected value remains unchanged relative to the dense baseline. For \ours{}, we retain the dense baseline's learning rate without additional tuning, to assess how closely approximate
rematerialization reproduces training with exact gradients
under the same optimization settings. All baselines are trained on the same number of tokens.

\paragraph{Compute environment.}
All the main experiments use BF16 mixed precision on 4 NVIDIA H100 or A100 GPUs, which improves speed without observable performance degradation. All experiments are implemented using the
TorchTitan~\cite{liang2025torchtitan} large-scale training framework. 

\subsection{Results}
\label{sec:main-results}

Tab.~\ref{tab:remat-results} presents the main results for \ours\,on models that are at most 488M parameters, for both token- and batch-level granularities. Vanilla denotes standard AC, while Top-, Opt-, and Rand-$\frac{k}{n}$ are
the selective variants described in Sec.~\ref{sec:main-method}. We report
FLOPs relative to vanilla backpropagation. These ratios provide a hardware-independent reference for the amount of arithmetic performed, but do not imply proportional wall-clock speedups, which also depend on the selection granularity and kernel efficiency discussed in App.~\ref{sec:main-block}. For example, a Llama-488M model with standard AC has an AC-FLOP ratio of $1.33$ due to the second forward recomputation. We evaluate \ours{} with
$\frac{k}{n}\in\left\{\frac{1}{3},\frac{2}{3}\right\}$. We further evaluate \ours{} with token-level Top-$\frac{2}{3}$ selection applied to FFN layers on a 953M-parameter Llama-style model using the same training protocol. We obtain a perplexity
of 9.959 compared to 9.858 for the dense baseline (a 1.02\% increase; App.~\ref{app:1b-experiment}). On the same model, batch-level FFN \ours{} reduces training-step
time by 18\% and 26\% with Top-$\frac{2}{3}$ and
Top-$\frac{1}{3}$ selection, respectively, relative to
Full AC on four NVIDIA H100 GPUs (App.~\ref{sec:main-block}).

\begin{figure}[t]
\centering
\begin{subfigure}[t]{0.48\linewidth}
    \centering
    \includegraphics[width=\linewidth]{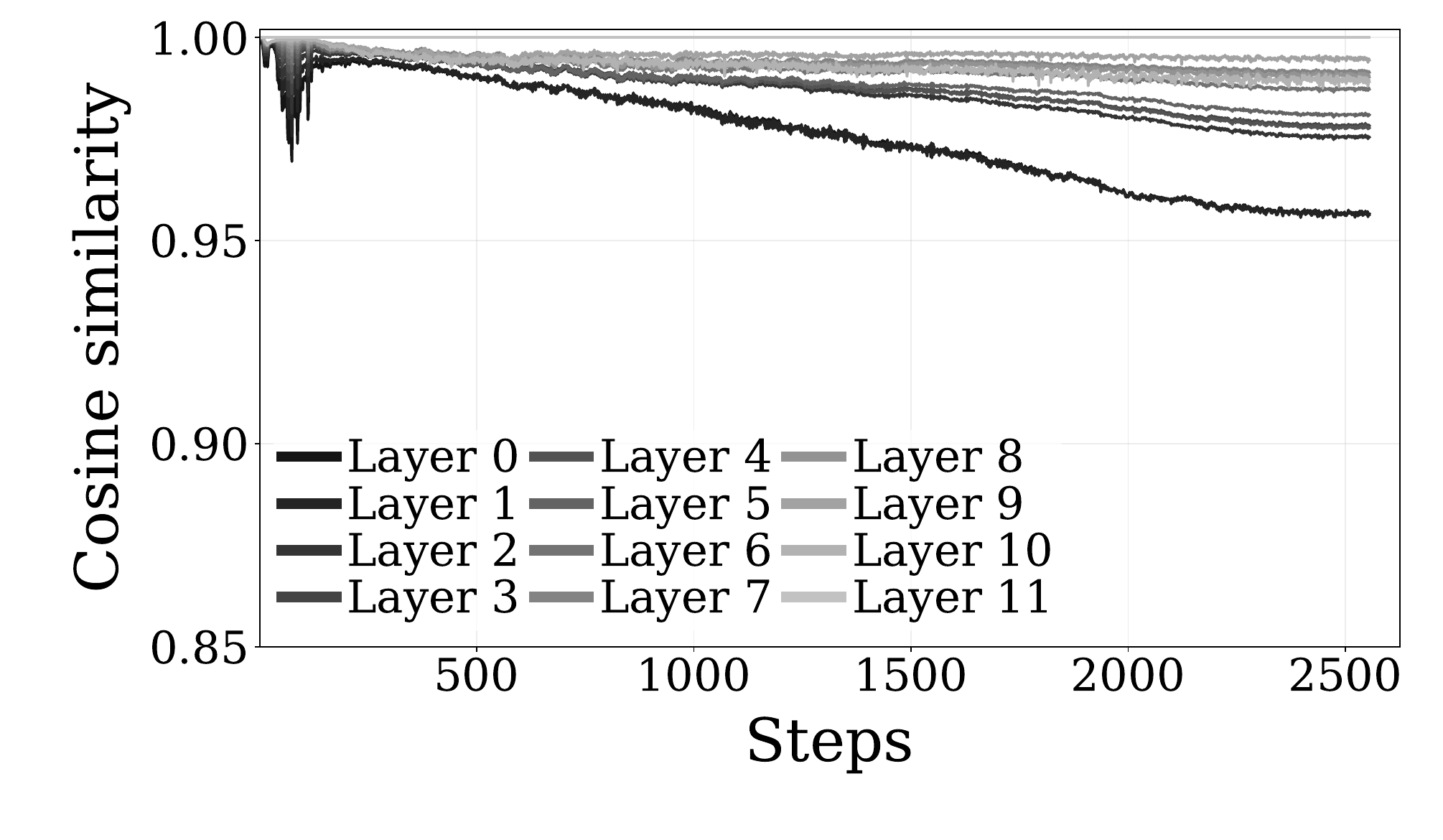}
    \caption{Attention}
    \label{fig:attn_cosine}
\end{subfigure}
\hfill
\begin{subfigure}[t]{0.48\linewidth}
    \centering
    \includegraphics[width=\linewidth]{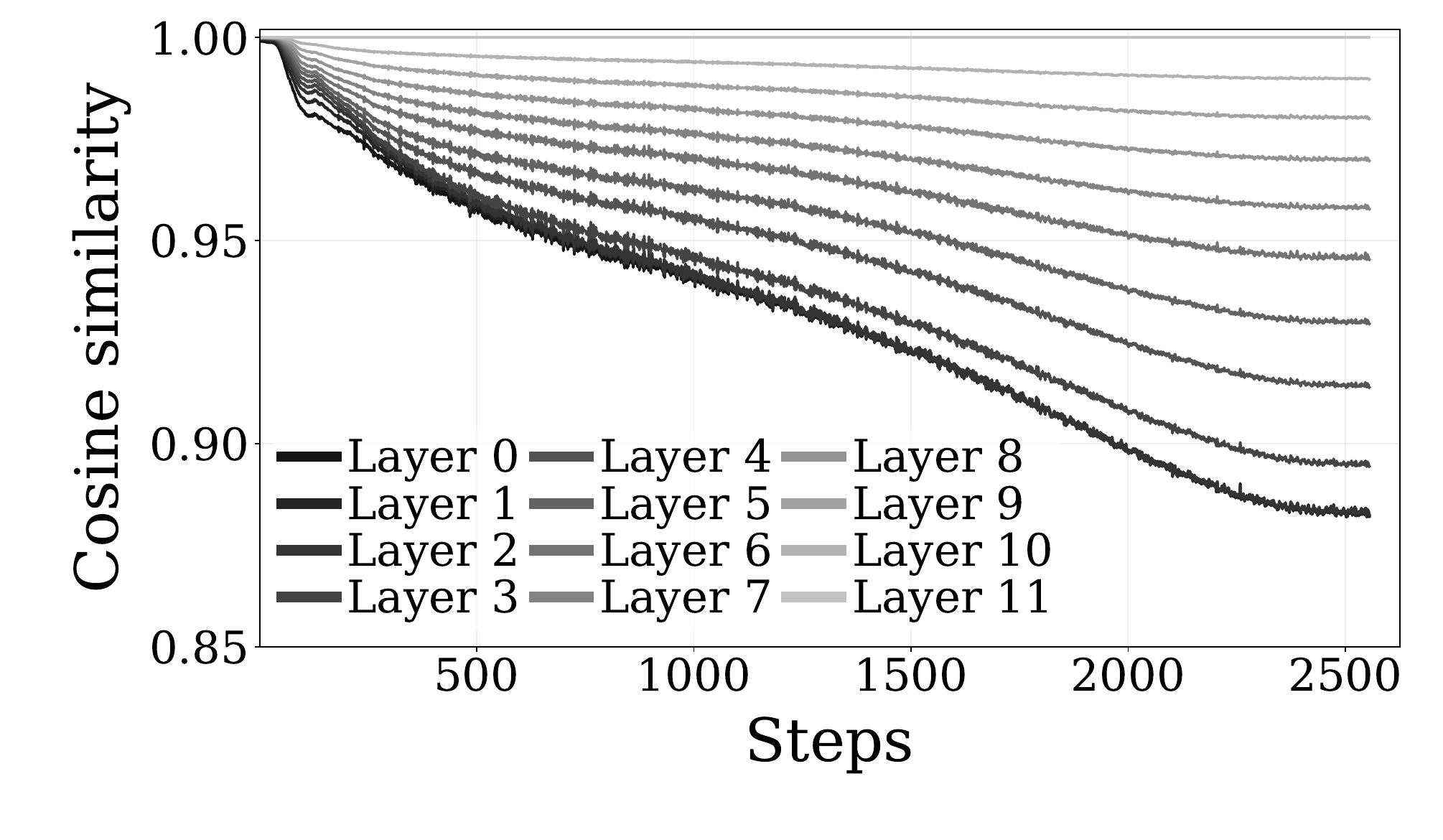}
    \caption{FFN}
    \label{fig:ffn_cosine}
\end{subfigure}
\caption{\textbf{Average cosine similarity between exact and \ours{} activation gradients} for a Llama-134M model $\Phi$, using \ours\, Top-$\frac{1}{3}$ with token-level granularity. These plots show that gradients remain well aligned across  layers, with only minor degradation from error propagation with depth.}
\label{fig:main-ablation_cosine}
\end{figure}

\paragraph{Rand-$\frac{k}{n}$ vs. Opt-$\frac{k}{n}$ vs. Top-$\frac{k}{n}$.}
As discussed in Sec.~\ref{sec:main-metod-kweak}, Opt-$\frac{k}{n}$ is constructed to minimize the mean-squared approximation error, and therefore consistently improves over Rand-$\frac{k}{n}$ method. Nevertheless, for a fixed $\frac{k}{n}$, our Top-$\frac{k}{n}$ approximation achieves substantially better performance than both baselines. Despite being biased, Top-$k$ selection rules are actually well known, for instance in gradient compression, to outperform random or variance-optimal baselines in practice \cite{beznosikov2023biased}.  In particular, relative to the standard Llama-488M, token-level Top-$\frac{2}{3}$ incurs at most a 2.5\% increase in perplexity in every setting.

\begin{wrapfigure}[16]{r}{0.5\textwidth}
  \centering
  \vspace{-20pt}
  \includegraphics[width=\linewidth]{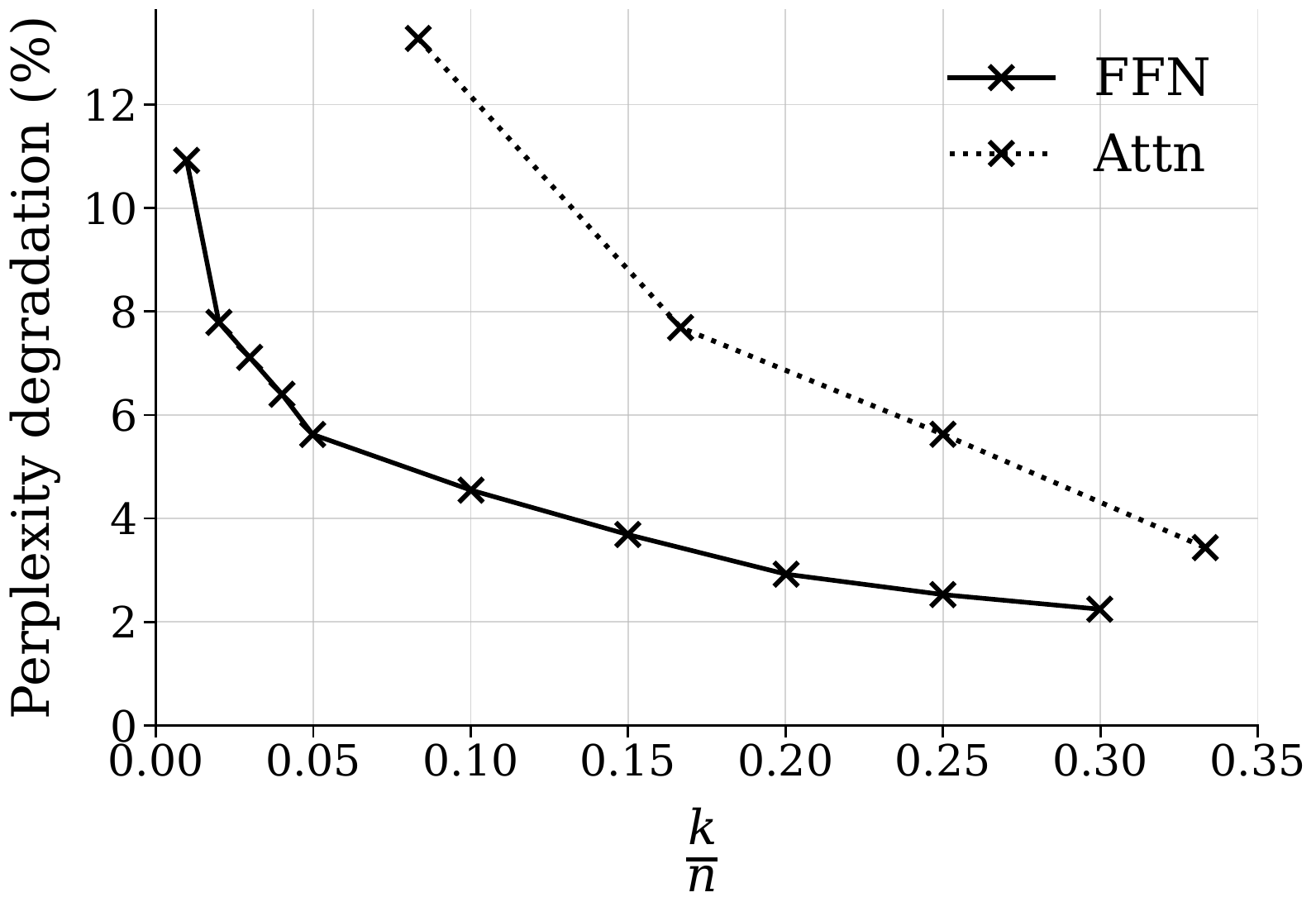}
  \caption{\textbf{Perplexity degradation versus $\frac{k}{n}$} for \ours{} Top-$\frac{k}{n}$ with token-level granularity applied on FFN and Attn for a Llama-134M model.}
  \label{fig:main-ablation_k}
\end{wrapfigure}
\paragraph{Smaller models vs. Top-$\frac{k}{n}$.} The smaller models provide a reference for allocating the same analytic FLOP budget to reduced model capacity rather than approximate rematerialization. At token-level granularity, compute-matched smaller models perform worse than or the same as \ours{} Top-$\frac{2}{3}$, suggesting that \ours{} applied to the original model can be preferable to reducing the corresponding FFN or attention dimensions in terms of accuracy. In particular, for attention, the accuracy of the Llama-468M model is the same as \ours. This is unsurprising: reducing attention removes few parameters and thus leaves little room for \ours{} to improve over it. For FFN, \ours{} outperforms the FLOP-matched smaller model. The ablations below further show that  Top-$\frac{k}{n}$ preserves a high-quality gradient approximation. Finally, batch-level granularity performs poorly, as discussed below.

\paragraph{Token-level vs. batch-level granularity.}Tab.~\ref{tab:token-level} compared to Tab.~\ref{tab:batch-level} shows that token-level gradient approximations consistently outperform batch-level ones, and batch-level approximations perform worse than a smaller model. This behavior is expected: as discussed in Sec.~\ref{sec:implem}, token-level granularity yields a more accurate gradient approximation in expectation at the cost of worse (but expected) hardware utilisation we discuss later in App.~\ref{sec:main-block}.

\paragraph{FFN vs. Attention Rematerialization.}
Applying \ours{} to FFN layers generally performs better than applying it to attention layers, or to both modules simultaneously. This is unsurprising: viewing the FFN as a one-hidden-layer network (Eq.~\eqref{eq:FFN}), the amplitudes of its gradient atoms are largely predictable from the forward pass. By contrast, this connection is less direct for attention, which has a more nonlinear gradient structure. Nevertheless, applying \ours{} jointly to FFN and attention layers preserves a useful approximation of the dense gradient while retaining a substantially lower activation-memory footprint than Selective AC.

\subsection{Ablation experiments}\label{sec:ablations}
We now ablate token-level \ours{} Top-$\frac{k}{n}$ on a smaller 134M Llama-style model.

\paragraph{Cosine similarity between approximate and exact gradients.}
In Fig.~\ref{fig:main-ablation_cosine}, we report the cosine similarity between the exact activation gradients and the \ours{} activation gradients for both FFN and attention layers, while training with the approximate gradient $\nabla \Phi^\omega$ from \ours. This allows us to verify that the resulting optimization dynamics not only achieve good performance, but also remain close to those of standard training. We use the Top-$\frac{k}{n}$ approximation with $\frac{k}{n}=\frac{1}{3}$ on all attention and FFN layers. Cosine similarities are computed at each layer, averaged over all tokens in a batch, and tracked throughout training. The fact that they remain above $0.85$ over the entire run indicates that the sparse reconstruction preserves the direction of the dense local gradient per step. This is notable, since $\omega$ is constructed solely from information available during the forward pass. Finally, we observe that cosine similarity tends to decrease with depth, likely due to a progressive degradation of the gradient approximation across layers. This behavior is expected and does not appear to hinder training.

\paragraph{Ablation on $\frac{k}{n}$.}
In Fig.~\ref{fig:main-ablation_k}, we investigate how performance degrades as the retained fraction $\frac{k}{n}$ varies, when applying \ours{} either exclusively to FFN layers or exclusively to attention layers. We terminate each sweep once the performance drop exceeds $10\%$. The results show that, due to the adaptivity of the selection rule, retaining only a small fraction of separable components can still preserve strong performance. The useful components appear highly concentrated: retaining only $5\%$ of the FFN coordinates still keeps the performance drop below $5\%$.%

\section{Conclusion}
In this work, we introduce \ours{}, a method for approximate rematerialization that exploits exact forward-pass checkpoints to preserve high-quality gradient estimates. Using information collected during the initial forward pass, \ours{} recomputes only a subset of the original FFN and attention components during rematerialization. On Llama-style models with up to 488M parameters, retaining two thirds of these components limits the perplexity increase to at most $2.5\%$ while maintaining the activation-memory benefits of full Activation Checkpointing. At token-level granularity, \ours{} also achieves comparable or better model quality than smaller FLOP-matched models, indicating that approximating rematerialization can preserve model capacity more effectively than reducing model size. Future work will focus on further optimizing finer-grained implementations to more fully translate the FLOP reductions into wall-clock speedups on modern accelerators.
\section*{Acknowledgements}
This work was supported by funding managed by the French National Research Agency (ANR) under the France 2030 program through the DAIMOS project (grant ANR-25-EXNU-0002) and  PEPR IA (grant
SHARP ANR-23-PEIA-0008). Part of this work was
granted access to the JeanZay HPC/AI resources of IDRIS under the allocation AD011015884R1, AD011017481, AD011016641, AD011017661 and AD011017766.


\bibliographystyle{plainnat}
\bibliography{references}

\begin{thebibliography}{56}
\providecommand{\natexlab}[1]{#1}
\providecommand{\url}[1]{\texttt{#1}}
\expandafter\ifx\csname urlstyle\endcsname\relax
  \providecommand{\doi}[1]{doi: #1}\else
  \providecommand{\doi}{doi: \begingroup \urlstyle{rm}\Url}\fi

\bibitem[Aguila-Multner et~al.(2025)Aguila-Multner, Beaumont, Eyraud-Dubois, and Gusak]{aguila2025optimized}
Adrien Aguila-Multner, Olivier Beaumont, Lionel Eyraud-Dubois, and Julia Gusak.
\newblock Optimized forward-backward rematerialization for memory-efficient pipeline parallel training, 2025.
\newblock URL \url{https://inria.hal.science/hal-05151601}.

\bibitem[Bakong et~al.(2026)Bakong, Massoulié, Oyallon, and Scaman]{bakong2026unbiased}
Killian Bakong, Laurent Massoulié, Edouard Oyallon, and Kevin Scaman.
\newblock Unbiased approximate vector-jacobian products for efficient backpropagation, 2026.
\newblock URL \url{https://arxiv.org/abs/2602.14701}.

\bibitem[Baydin et~al.(2018)Baydin, Pearlmutter, Radul, and Siskind]{baydin2018automatic}
Atilim~Gunes Baydin, Barak~A Pearlmutter, Alexey~Andreyevich Radul, and Jeffrey~Mark Siskind.
\newblock Automatic differentiation in machine learning: a survey.
\newblock \emph{Journal of machine learning research}, 18\penalty0 (153):\penalty0 1--43, 2018.

\bibitem[Beaumont et~al.(2024)Beaumont, Eyraud-Dubois, Herrmann, Joly, and Shilova]{beaumont2024optimal}
Olivier Beaumont, Lionel Eyraud-Dubois, Julien Herrmann, Alexis Joly, and Alena Shilova.
\newblock Optimal re-materialization strategies for heterogeneous chains: How to train deep neural networks with limited memory.
\newblock \emph{ACM Trans. Math. Softw.}, 50\penalty0 (2), June 2024.
\newblock ISSN 0098-3500.
\newblock \doi{10.1145/3648633}.
\newblock URL \url{https://doi.org/10.1145/3648633}.

\bibitem[Belilovsky et~al.(2020)Belilovsky, Eickenberg, and Oyallon]{belilovsky2020decoupled}
Eugene Belilovsky, Michael Eickenberg, and Edouard Oyallon.
\newblock Decoupled greedy learning of {CNNs}.
\newblock In \emph{Proceedings of the 37th International Conference on Machine Learning}, ICML'20. JMLR.org, 2020.

\bibitem[Beznosikov et~al.(2023)Beznosikov, Horv{\'a}th, Richt{\'a}rik, and Safaryan]{beznosikov2023biased}
Aleksandr Beznosikov, Samuel Horv{\'a}th, Peter Richt{\'a}rik, and Mher Safaryan.
\newblock On biased compression for distributed learning.
\newblock \emph{Journal of Machine Learning Research}, 24\penalty0 (276):\penalty0 1--50, 2023.
\newblock URL \url{http://jmlr.org/papers/v24/21-1548.html}.

\bibitem[Chakrabarti and Moseley(2019)]{chakrabarti2019backprop}
Ayan Chakrabarti and Benjamin Moseley.
\newblock Backprop with approximate activations for memory-efficient network training.
\newblock In H.~Wallach, H.~Larochelle, A.~Beygelzimer, F.~d\textquotesingle Alch\'{e}-Buc, E.~Fox, and R.~Garnett, editors, \emph{Advances in Neural Information Processing Systems}, volume~32. Curran Associates, Inc., 2019.
\newblock URL \url{https://proceedings.neurips.cc/paper_files/paper/2019/file/30c8e1ca872524fbf7ea5c519ca397ee-Paper.pdf}.

\bibitem[Chen et~al.(2025)Chen, Deng, Li, He, Zhu, Zheng, Wang, Huai, and Guo]{chen2025adacc}
Ping Chen, Zhuohong Deng, Ping Li, Shuibing He, Hongzi Zhu, Yi~Zheng, Zhefeng Wang, Baoxing Huai, and Minyi Guo.
\newblock Adacc: An adaptive framework unifying compression and activation recomputation for {LLM} training, 2025.
\newblock URL \url{https://arxiv.org/abs/2508.00806}.

\bibitem[Chmiel et~al.(2023)Chmiel, Hubara, Banner, and Soudry]{chmiel2023mvue}
Brian Chmiel, Itay Hubara, Ron Banner, and Daniel Soudry.
\newblock Minimum variance unbiased {N:M} sparsity for the neural gradients.
\newblock In \emph{The Eleventh International Conference on Learning Representations}, 2023.
\newblock URL \url{https://openreview.net/forum?id=vuD2xEtxZcj}.

\bibitem[Daniely et~al.(2017)Daniely, Lazic, Singer, and Talwar]{daniely2017short}
Amit Daniely, Nevena Lazic, Yoram Singer, and Kunal Talwar.
\newblock Short and deep: Sketching and neural networks.
\newblock In \emph{International Conference on Learning Representations Workshop}, 2017.
\newblock URL \url{https://openreview.net/pdf?id=r1br_2Kge}.

\bibitem[Dao et~al.(2022)Dao, Fu, Ermon, Rudra, and R{\'e}]{dao2022flashattention}
Tri Dao, Daniel~Y. Fu, Stefano Ermon, Atri Rudra, and Christopher R{\'e}.
\newblock {FlashAttention}: fast and memory-efficient exact attention with {IO}-awareness.
\newblock In \emph{Proceedings of the 36th International Conference on Neural Information Processing Systems}, NIPS '22, Red Hook, NY, USA, 2022. Curran Associates Inc.
\newblock ISBN 9781713871088.

\bibitem[DeepSeek-AI et~al.(2025)DeepSeek-AI, Liu, Feng, Xue, Wang, Wu, Lu, Zhao, Deng, Zhang, Ruan, et~al.]{liu2024deepseek}
DeepSeek-AI, Aixin Liu, Bei Feng, Bing Xue, Bingxuan Wang, Bochao Wu, Chengda Lu, Chenggang Zhao, Chengqi Deng, Chenyu Zhang, Chong Ruan, et~al.
\newblock {DeepSeek-V3} technical report, 2025.
\newblock URL \url{https://arxiv.org/abs/2412.19437}.

\bibitem[{Epoch AI}(2026)]{epochai2026parametertrends}
{Epoch AI}.
\newblock Exponential growth of parameters in notable {AI} systems, 2026.
\newblock URL \url{https://ourworldindata.org/grapher/exponential-growth-of-parameters-in-notable-ai-systems}.

\bibitem[Fournier et~al.(2024)Fournier, Nabli, Aminbeidokhti, Pedersoli, Belilovsky, and Oyallon]{fournier2024wash}
Louis Fournier, Adel Nabli, Masih Aminbeidokhti, Marco Pedersoli, Eugene Belilovsky, and Edouard Oyallon.
\newblock {WASH}: Train your ensemble with communication-efficient weight shuffling, then average.
\newblock In \emph{OPT 2024: Optimization for Machine Learning}, 2024.

\bibitem[Gazagnadou et~al.(2022)Gazagnadou, Ibrahim, and Gower]{gazagnadou2022ridgesketch}
Nidham Gazagnadou, Mark Ibrahim, and Robert~M. Gower.
\newblock {RidgeSketch}: A fast sketching based solver for large scale ridge regression.
\newblock \emph{SIAM Journal on Matrix Analysis and Applications}, 43\penalty0 (3):\penalty0 1440--1468, 2022.
\newblock \doi{10.1137/21M1422963}.
\newblock URL \url{https://doi.org/10.1137/21M1422963}.

\bibitem[Gomez et~al.(2017)Gomez, Ren, Urtasun, and Grosse]{gomez2017reversible}
Aidan~N Gomez, Mengye Ren, Raquel Urtasun, and Roger~B Grosse.
\newblock The reversible residual network: Backpropagation without storing activations.
\newblock In I.~Guyon, U.~Von Luxburg, S.~Bengio, H.~Wallach, R.~Fergus, S.~Vishwanathan, and R.~Garnett, editors, \emph{Advances in Neural Information Processing Systems}, volume~30. Curran Associates, Inc., 2017.
\newblock URL \url{https://proceedings.neurips.cc/paper_files/paper/2017/file/f9be311e65d81a9ad8150a60844bb94c-Paper.pdf}.

\bibitem[Grattafiori et~al.(2024)Grattafiori, Dubey, Jauhri, Pandey, Kadian, Al-Dahle, Letman, Mathur, Schelten, Vaughan, et~al.]{grattafiori2024llama}
Aaron Grattafiori, Abhimanyu Dubey, Abhinav Jauhri, Abhinav Pandey, Abhishek Kadian, Ahmad Al-Dahle, Aiesha Letman, Akhil Mathur, Alan Schelten, Alex Vaughan, et~al.
\newblock The {Llama} 3 herd of models.
\newblock \emph{arXiv preprint arXiv:2407.21783}, 2024.

\bibitem[Gray et~al.(2017)Gray, Radford, and Kingma]{gray2017blocksparse}
Scott Gray, Alec Radford, and Diederik~P. Kingma.
\newblock {GPU} kernels for block-sparse weights.
\newblock Technical report, OpenAI, 2017.
\newblock URL \url{https://cdn.openai.com/blocksparse/blocksparsepaper.pdf}.

\bibitem[Griewank(1992)]{griewank1992achieving}
Andreas Griewank.
\newblock Achieving logarithmic growth of temporal and spatial complexity in reverse automatic differentiation.
\newblock \emph{Optimization Methods and Software}, 1\penalty0 (1):\penalty0 35--54, 1992.
\newblock \doi{10.1080/10556789208805505}.
\newblock URL \url{https://doi.org/10.1080/10556789208805505}.

\bibitem[Griewank and Walther(2000)]{griewank2000algorithm}
Andreas Griewank and Andrea Walther.
\newblock Algorithm 799: revolve: an implementation of checkpointing for the reverse or adjoint mode of computational differentiation.
\newblock \emph{ACM Trans. Math. Softw.}, 26\penalty0 (1):\penalty0 19–45, March 2000.
\newblock ISSN 0098-3500.
\newblock \doi{10.1145/347837.347846}.
\newblock URL \url{https://doi.org/10.1145/347837.347846}.

\bibitem[Gusak et~al.(2025)Gusak, Zhao, Le~Hellard, Li, Eyraud-Dubois, and Beaumont]{gusak2025hiremate}
Julia Gusak, Xunyi Zhao, Th\'{e}otime Le~Hellard, Zhe Li, Lionel Eyraud-Dubois, and Olivier Beaumont.
\newblock {H}i{R}emate: Hierarchical approach for efficient re-materialization of neural networks.
\newblock In Aarti Singh, Maryam Fazel, Daniel Hsu, Simon Lacoste-Julien, Felix Berkenkamp, Tegan Maharaj, Kiri Wagstaff, and Jerry Zhu, editors, \emph{Proceedings of the 42nd International Conference on Machine Learning}, volume 267 of \emph{Proceedings of Machine Learning Research}, pages 21418--21443. PMLR, 13--19 Jul 2025.
\newblock URL \url{https://proceedings.mlr.press/v267/gusak25a.html}.

\bibitem[Hoffmann et~al.(2022)Hoffmann, Borgeaud, Mensch, Buchatskaya, Cai, Rutherford, de~Las~Casas, Hendricks, Welbl, Clark, Hennigan, Noland, Millican, van~den Driessche, Damoc, Guy, Osindero, Simonyan, Elsen, Vinyals, Rae, and Sifre]{hoffmann2022training}
Jordan Hoffmann, Sebastian Borgeaud, Arthur Mensch, Elena Buchatskaya, Trevor Cai, Eliza Rutherford, Diego de~Las~Casas, Lisa~Anne Hendricks, Johannes Welbl, Aidan Clark, Thomas Hennigan, Eric Noland, Katherine Millican, George van~den Driessche, Bogdan Damoc, Aurelia Guy, Simon Osindero, Kar\'{e}n Simonyan, Erich Elsen, Oriol Vinyals, Jack Rae, and Laurent Sifre.
\newblock An empirical analysis of compute-optimal large language model training.
\newblock In S.~Koyejo, S.~Mohamed, A.~Agarwal, D.~Belgrave, K.~Cho, and A.~Oh, editors, \emph{Advances in Neural Information Processing Systems}, volume~35, pages 30016--30030. Curran Associates, Inc., 2022.
\newblock \doi{10.52202/068431-2176}.
\newblock URL \url{https://proceedings.neurips.cc/paper_files/paper/2022/file/c1e2faff6f588870935f114ebe04a3e5-Paper-Conference.pdf}.

\bibitem[Huang et~al.(2019)Huang, Cheng, Bapna, Firat, Chen, Chen, Lee, Ngiam, Le, Wu, and Chen]{huang2019gpipe}
Yanping Huang, Youlong Cheng, Ankur Bapna, Orhan Firat, Dehao Chen, Mia Chen, HyoukJoong Lee, Jiquan Ngiam, Quoc~V Le, Yonghui Wu, and Zhifeng Chen.
\newblock {GPipe}: Efficient training of giant neural networks using pipeline parallelism.
\newblock In H.~Wallach, H.~Larochelle, A.~Beygelzimer, F.~d\textquotesingle Alch\'{e}-Buc, E.~Fox, and R.~Garnett, editors, \emph{Advances in Neural Information Processing Systems}, volume~32. Curran Associates, Inc., 2019.
\newblock URL \url{https://proceedings.neurips.cc/paper_files/paper/2019/file/093f65e080a295f8076b1c5722a46aa2-Paper.pdf}.

\bibitem[Jacobsen et~al.(2018)Jacobsen, Smeulders, and Oyallon]{jacobsen2018revnet}
Jörn-Henrik Jacobsen, Arnold~W.M. Smeulders, and Edouard Oyallon.
\newblock {i-RevNet}: Deep invertible networks.
\newblock In \emph{International Conference on Learning Representations}, 2018.
\newblock URL \url{https://openreview.net/forum?id=HJsjkMb0Z}.

\bibitem[Janson et~al.(2026{\natexlab{a}})Janson, Oyallon, and Belilovsky]{janson2026learned}
Paul Janson, Edouard Oyallon, and Eugene Belilovsky.
\newblock Learned subspace compression for communication-efficient pipeline parallelism.
\newblock \emph{arXiv preprint arXiv:2606.05484}, 2026{\natexlab{a}}.

\bibitem[Janson et~al.(2026{\natexlab{b}})Janson, Oyallon, and Belilovsky]{janson2026stabilizing}
Paul Janson, Edouard Oyallon, and Eugene Belilovsky.
\newblock Stabilizing native low-rank {LLM} pretraining.
\newblock In \emph{Proceedings of the 43rd International Conference on Machine Learning}, volume 306 of \emph{Proceedings of Machine Learning Research}, pages 50830--50852. PMLR, 2026{\natexlab{b}}.
\newblock URL \url{https://proceedings.mlr.press/v306/janson26a.html}.

\bibitem[Jordan et~al.(2024)Jordan, Jin, Boza, You, Cesista, Newhouse, and Bernstein]{keller2024muon}
Keller Jordan, Yuchen Jin, Vlado Boza, Jiacheng You, Franz Cesista, Laker Newhouse, and Jeremy Bernstein.
\newblock Muon: An optimizer for hidden layers in neural networks, 2024.
\newblock URL \url{https://kellerjordan.github.io/posts/muon/}.

\bibitem[Kim et~al.(2023)Kim, Kim, Yu, and Chun]{kim2023bpipe}
Taebum Kim, Hyoungjoo Kim, Gyeong-In Yu, and Byung-Gon Chun.
\newblock {BP}ipe: Memory-balanced pipeline parallelism for training large language models.
\newblock In Andreas Krause, Emma Brunskill, Kyunghyun Cho, Barbara Engelhardt, Sivan Sabato, and Jonathan Scarlett, editors, \emph{Proceedings of the 40th International Conference on Machine Learning}, volume 202 of \emph{Proceedings of Machine Learning Research}, pages 16639--16653. PMLR, 23--29 Jul 2023.
\newblock URL \url{https://proceedings.mlr.press/v202/kim23l.html}.

\bibitem[Kong et~al.(2026)Kong, Huang, Xu, Liang, Wang, and Yuan]{kong2025clapping}
Boao Kong, Xu~Huang, Yuqi Xu, Yixuan Liang, Bin Wang, and Kun Yuan.
\newblock Clapping: Removing per-sample storage for pipeline parallel distributed optimization with communication compression, 2026.
\newblock URL \url{https://arxiv.org/abs/2509.19029}.

\bibitem[Korthikanti et~al.(2023)Korthikanti, Casper, Lym, McAfee, Andersch, Shoeybi, and Catanzaro]{korthikanti2023reducing}
Vijay~Anand Korthikanti, Jared Casper, Sangkug Lym, Lawrence McAfee, Michael Andersch, Mohammad Shoeybi, and Bryan Catanzaro.
\newblock Reducing activation recomputation in large transformer models.
\newblock In D.~Song, M.~Carbin, and T.~Chen, editors, \emph{Proceedings of Machine Learning and Systems}, volume~5, pages 341--353. Curran Associates, Inc., 2023.
\newblock URL \url{https://proceedings.mlsys.org/paper_files/paper/2023/file/80083951326cf5b35e5100260d64ed81-Paper-mlsys2023.pdf}.

\bibitem[Liang et~al.(2025)Liang, Liu, Wright, Constable, Gu, Huang, Zhang, Feng, Huang, Wang, Purandare, Nadathur, and Idreos]{liang2025torchtitan}
Wanchao Liang, Tianyu Liu, Less Wright, Will Constable, Andrew Gu, Chien-Chin Huang, Iris Zhang, Wei Feng, Howard Huang, Junjie Wang, Sanket Purandare, Gokul Nadathur, and Stratos Idreos.
\newblock {TorchTitan}: One-stop {PyTorch} native solution for production ready {LLM} pretraining.
\newblock In \emph{The Thirteenth International Conference on Learning Representations}, 2025.
\newblock URL \url{https://openreview.net/forum?id=SFN6Wm7YBI}.

\bibitem[Linnainmaa(1976)]{linnainmaa1976taylor}
Seppo Linnainmaa.
\newblock Taylor expansion of the accumulated rounding error.
\newblock \emph{BIT}, 16\penalty0 (2):\penalty0 146–160, June 1976.
\newblock ISSN 0006-3835.
\newblock \doi{10.1007/BF01931367}.
\newblock URL \url{https://doi.org/10.1007/BF01931367}.

\bibitem[Liu et~al.(2022)Liu, Zheng, Wang, Cen, Chen, Han, Chen, Liu, Tang, Gonzalez, Mahoney, and Cheung]{liu2022gact}
Xiaoxuan Liu, Lianmin Zheng, Dequan Wang, Yukuo Cen, Weize Chen, Xu~Han, Jianfei Chen, Zhiyuan Liu, Jie Tang, Joey Gonzalez, Michael Mahoney, and Alvin Cheung.
\newblock {GACT}: Activation compressed training for generic network architectures.
\newblock In Kamalika Chaudhuri, Stefanie Jegelka, Le~Song, Csaba Szepesvari, Gang Niu, and Sivan Sabato, editors, \emph{Proceedings of the 39th International Conference on Machine Learning}, volume 162 of \emph{Proceedings of Machine Learning Research}, pages 14139--14152. PMLR, 17--23 Jul 2022.
\newblock URL \url{https://proceedings.mlr.press/v162/liu22v.html}.

\bibitem[Loshchilov and Hutter(2017)]{loshchilov2017sgdr}
Ilya Loshchilov and Frank Hutter.
\newblock {SGDR}: Stochastic gradient descent with warm restarts.
\newblock In \emph{International Conference on Learning Representations}, 2017.
\newblock URL \url{https://openreview.net/forum?id=Skq89Scxx}.

\bibitem[Loshchilov and Hutter(2019)]{loshchilov2019decoupled}
Ilya Loshchilov and Frank Hutter.
\newblock Decoupled weight decay regularization.
\newblock In \emph{International Conference on Learning Representations}, 2019.
\newblock URL \url{https://openreview.net/forum?id=Bkg6RiCqY7}.

\bibitem[Micikevicius et~al.(2018)Micikevicius, Narang, Alben, Diamos, Elsen, Garcia, Ginsburg, Houston, Kuchaiev, Venkatesh, and Wu]{micikevicius2018mixed}
Paulius Micikevicius, Sharan Narang, Jonah Alben, Gregory Diamos, Erich Elsen, David Garcia, Boris Ginsburg, Michael Houston, Oleksii Kuchaiev, Ganesh Venkatesh, and Hao Wu.
\newblock Mixed precision training.
\newblock In \emph{International Conference on Learning Representations}, 2018.
\newblock URL \url{https://openreview.net/forum?id=r1gs9JgRZ}.

\bibitem[Nabli et~al.(2025)Nabli, Fournier, ERBACHER, Serrano, Belilovsky, and Oyallon]{nabli2026acco}
Adel Nabli, Louis Fournier, Pierre ERBACHER, Louis Serrano, Eugene Belilovsky, and Edouard Oyallon.
\newblock {ACCO}: Accumulate while you communicate for communication-overlapped sharded {LLM} training.
\newblock In D.~Belgrave, C.~Zhang, H.~Lin, R.~Pascanu, P.~Koniusz, M.~Ghassemi, and N.~Chen, editors, \emph{Advances in Neural Information Processing Systems}, volume 38, Main Conference, pages 69278--69305. Curran Associates, Inc., 2025.
\newblock \doi{10.52202/085713-2329}.
\newblock URL \url{https://proceedings.neurips.cc/paper_files/paper/2025/file/6454dcd80b5373daaa97e53ce32c78a1-Paper-Conference.pdf}.

\bibitem[Paszke et~al.(2019)Paszke, Gross, Massa, Lerer, Bradbury, Chanan, Killeen, Lin, Gimelshein, Antiga, Desmaison, Kopf, Yang, DeVito, Raison, Tejani, Chilamkurthy, Steiner, Fang, Bai, and Chintala]{paszke2019pytorch}
Adam Paszke, Sam Gross, Francisco Massa, Adam Lerer, James Bradbury, Gregory Chanan, Trevor Killeen, Zeming Lin, Natalia Gimelshein, Luca Antiga, Alban Desmaison, Andreas Kopf, Edward Yang, Zachary DeVito, Martin Raison, Alykhan Tejani, Sasank Chilamkurthy, Benoit Steiner, Lu~Fang, Junjie Bai, and Soumith Chintala.
\newblock {PyTorch}: An imperative style, high-performance deep learning library.
\newblock In H.~Wallach, H.~Larochelle, A.~Beygelzimer, F.~d\textquotesingle Alch\'{e}-Buc, E.~Fox, and R.~Garnett, editors, \emph{Advances in Neural Information Processing Systems}, volume~32. Curran Associates, Inc., 2019.
\newblock URL \url{https://proceedings.neurips.cc/paper_files/paper/2019/file/bdbca288fee7f92f2bfa9f7012727740-Paper.pdf}.

\bibitem[Penedo et~al.(2024)Penedo, Kydl\'{\i}\v{c}ek, Ben~Allal, Lozhkov, Mitchell, Raffel, Von~Werra, and Wolf]{penedo2024fineweb}
Guilherme Penedo, Hynek Kydl\'{\i}\v{c}ek, Loubna Ben~Allal, Anton Lozhkov, Margaret Mitchell, Colin Raffel, Leandro Von~Werra, and Thomas Wolf.
\newblock The {FineWeb} datasets: Decanting the web for the finest text data at scale.
\newblock In A.~Globerson, L.~Mackey, D.~Belgrave, A.~Fan, U.~Paquet, J.~Tomczak, and C.~Zhang, editors, \emph{Advances in Neural Information Processing Systems}, volume~37, pages 30811--30849. Curran Associates, Inc., 2024.
\newblock \doi{10.52202/079017-0970}.
\newblock URL \url{https://proceedings.neurips.cc/paper_files/paper/2024/file/370df50ccfdf8bde18f8f9c2d9151bda-Paper-Datasets_and_Benchmarks_Track.pdf}.

\bibitem[{PyTorch Foundation}(2025)]{pytorch_activation_checkpointing_2025}
{PyTorch Foundation}.
\newblock Current and new activation checkpointing techniques in {PyTorch}, March 2025.
\newblock URL \url{https://pytorch.org/blog/activation-checkpointing-techniques/}.

\bibitem[Rajbhandari et~al.(2020)Rajbhandari, Rasley, Ruwase, and He]{rajbhandari2020zero}
Samyam Rajbhandari, Jeff Rasley, Olatunji Ruwase, and Yuxiong He.
\newblock {ZeRO}: memory optimizations toward training trillion parameter models.
\newblock In \emph{Proceedings of the International Conference for High Performance Computing, Networking, Storage and Analysis}, SC '20. IEEE Press, 2020.
\newblock ISBN 9781728199986.

\bibitem[Ren et~al.(2021)Ren, Rajbhandari, Aminabadi, Ruwase, Yang, Zhang, Li, and He]{ren2021zero}
Jie Ren, Samyam Rajbhandari, Reza~Yazdani Aminabadi, Olatunji Ruwase, Shuangyan Yang, Minjia Zhang, Dong Li, and Yuxiong He.
\newblock {ZeRO-Offload}: Democratizing {Billion-Scale} model training.
\newblock In \emph{2021 USENIX Annual Technical Conference (USENIX ATC 21)}, pages 551--564. USENIX Association, July 2021.
\newblock ISBN 978-1-939133-23-6.
\newblock URL \url{https://www.usenix.org/conference/atc21/presentation/ren-jie}.

\bibitem[Rhu et~al.(2016)Rhu, Gimelshein, Clemons, Zulfiqar, and Keckler]{rhu2016vdnn}
Minsoo Rhu, Natalia Gimelshein, Jason Clemons, Arslan Zulfiqar, and Stephen~W. Keckler.
\newblock {vDNN}: virtualized deep neural networks for scalable, memory-efficient neural network design.
\newblock In \emph{The 49th Annual IEEE/ACM International Symposium on Microarchitecture}, MICRO-49. IEEE Press, 2016.

\bibitem[Rivaud et~al.(2025)Rivaud, Fournier, Pumir, Belilovsky, Eickenberg, and Oyallon]{rivaud2025petra}
Stephane Rivaud, Louis Fournier, Thomas Pumir, Eugene Belilovsky, Michael Eickenberg, and Edouard Oyallon.
\newblock {PETRA}: Parallel end-to-end training with reversible architectures.
\newblock In \emph{The Thirteenth International Conference on Learning Representations}, 2025.
\newblock URL \url{https://openreview.net/forum?id=0fhzSFsGUT}.

\bibitem[Sarfi et~al.(2025)Sarfi, Th{\'e}rien, Lidin, and Belilovsky]{sarfi2025communication}
Amir Sarfi, Benjamin Th{\'e}rien, Joel Lidin, and Eugene Belilovsky.
\newblock Communication efficient {LLM} pre-training with {SparseLoCo}.
\newblock \emph{arXiv preprint arXiv:2508.15706}, 2025.

\bibitem[Shamshoum et~al.(2025)Shamshoum, Hodos, Sieradzki, and Schuster]{shamshoum2025compact}
Yara Shamshoum, Nitzan Hodos, Yuval Sieradzki, and Assaf Schuster.
\newblock {C}omp{A}ct: Compressed activations for memory-efficient {LLM} training.
\newblock In Luis Chiruzzo, Alan Ritter, and Lu~Wang, editors, \emph{Proceedings of the 2025 Conference of the Nations of the Americas Chapter of the Association for Computational Linguistics: Human Language Technologies (Volume 1: Long Papers)}, pages 1511--1524, Albuquerque, New Mexico, April 2025. Association for Computational Linguistics.
\newblock ISBN 979-8-89176-189-6.
\newblock \doi{10.18653/v1/2025.naacl-long.71}.
\newblock URL \url{https://aclanthology.org/2025.naacl-long.71/}.

\bibitem[Singh et~al.(2025)Singh, Khalid, Cagnasso, Oyallon, and Belilovsky]{singh2025model}
Vaibhav Singh, Zafir Khalid, Pietro Cagnasso, Edouard Oyallon, and Eugene Belilovsky.
\newblock Model parallelism with subnetwork data parallelism.
\newblock \emph{arXiv preprint arXiv:2507.09029}, 2025.

\bibitem[Sozzo et~al.(2026)Sozzo, Fleming, Flamm, and Thompson]{delsozzo2026progress}
Emanuele~Del Sozzo, Martin Fleming, Kenneth Flamm, and Neil Thompson.
\newblock How much progress has there been in {NVIDIA} datacenter {GPU}s?, 2026.
\newblock URL \url{https://arxiv.org/abs/2601.20115}.

\bibitem[Su et~al.(2024)Su, Ahmed, Lu, Pan, Bo, and Liu]{su2024roformer}
Jianlin Su, Murtadha Ahmed, Yu~Lu, Shengfeng Pan, Wen Bo, and Yunfeng Liu.
\newblock {RoFormer}: Enhanced transformer with rotary position embedding.
\newblock \emph{Neurocomput.}, 568\penalty0 (C), February 2024.
\newblock ISSN 0925-2312.
\newblock \doi{10.1016/j.neucom.2023.127063}.
\newblock URL \url{https://doi.org/10.1016/j.neucom.2023.127063}.

\bibitem[Sun et~al.(2017)Sun, Ren, Ma, and Wang]{sun2017meprop}
Xu~Sun, Xuancheng Ren, Shuming Ma, and Houfeng Wang.
\newblock me{P}rop: Sparsified back propagation for accelerated deep learning with reduced overfitting.
\newblock In Doina Precup and Yee~Whye Teh, editors, \emph{Proceedings of the 34th International Conference on Machine Learning}, volume~70 of \emph{Proceedings of Machine Learning Research}, pages 3299--3308. PMLR, 06--11 Aug 2017.
\newblock URL \url{https://proceedings.mlr.press/v70/sun17c.html}.

\bibitem[Tay et~al.(2022)Tay, Dehghani, Rao, Fedus, Abnar, Chung, Narang, Yogatama, Vaswani, and Metzler]{tay2022scale}
Yi~Tay, Mostafa Dehghani, Jinfeng Rao, William Fedus, Samira Abnar, Hyung~Won Chung, Sharan Narang, Dani Yogatama, Ashish Vaswani, and Donald Metzler.
\newblock Scale efficiently: Insights from pretraining and finetuning transformers.
\newblock In \emph{International Conference on Learning Representations}, 2022.
\newblock URL \url{https://openreview.net/forum?id=f2OYVDyfIB}.

\bibitem[Vaswani et~al.(2017)Vaswani, Shazeer, Parmar, Uszkoreit, Jones, Gomez, Kaiser, and Polosukhin]{vaswani2017attention}
Ashish Vaswani, Noam Shazeer, Niki Parmar, Jakob Uszkoreit, Llion Jones, Aidan~N Gomez, {\L}ukasz Kaiser, and Illia Polosukhin.
\newblock Attention is all you need.
\newblock In I.~Guyon, U.~Von Luxburg, S.~Bengio, H.~Wallach, R.~Fergus, S.~Vishwanathan, and R.~Garnett, editors, \emph{Advances in Neural Information Processing Systems}, volume~30. Curran Associates, Inc., 2017.
\newblock URL \url{https://proceedings.neurips.cc/paper_files/paper/2017/file/3f5ee243547dee91fbd053c1c4a845aa-Paper.pdf}.

\bibitem[Woodruff(2014)]{woodruff2014sketching}
David~P. Woodruff.
\newblock Sketching as a tool for numerical linear algebra.
\newblock \emph{Found. Trends Theor. Comput. Sci.}, 10\penalty0 (1–2):\penalty0 1–157, October 2014.
\newblock ISSN 1551-305X.
\newblock \doi{10.1561/0400000060}.
\newblock URL \url{https://doi.org/10.1561/0400000060}.

\bibitem[Xu et~al.(2026)Xu, Ping, Xu, Liu, Wang, Shoeybi, Li, and Catanzaro]{xu2026128k}
Chejian Xu, Wei Ping, Peng Xu, Zihan Liu, Boxin Wang, Mohammad Shoeybi, Bo~Li, and Bryan Catanzaro.
\newblock From {128K} to {4M}: Efficient training of ultra-long context large language models.
\newblock In \emph{Findings of the Association for Computational Linguistics: ACL 2026}, pages 13122--13133, 2026.

\bibitem[Yang et~al.(2021)Yang, Hu, Babuschkin, Sidor, Liu, Farhi, Ryder, Pachocki, Chen, and Gao]{yang2021tuning}
Ge~Yang, Edward Hu, Igor Babuschkin, Szymon Sidor, Xiaodong Liu, David Farhi, Nick Ryder, Jakub Pachocki, Weizhu Chen, and Jianfeng Gao.
\newblock Tuning large neural networks via zero-shot hyperparameter transfer.
\newblock In M.~Ranzato, A.~Beygelzimer, Y.~Dauphin, P.S. Liang, and J.~Wortman Vaughan, editors, \emph{Advances in Neural Information Processing Systems}, volume~34, pages 17084--17097. Curran Associates, Inc., 2021.
\newblock URL \url{https://proceedings.neurips.cc/paper_files/paper/2021/file/8df7c2e3c3c3be098ef7b382bd2c37ba-Paper.pdf}.

\bibitem[Yang et~al.(2022)Yang, Xu, Wen, Chen, and Xu]{yang2022sketch}
Minghan Yang, Dong Xu, Zaiwen Wen, Mengyun Chen, and Pengxiang Xu.
\newblock Sketch-based empirical natural gradient methods for deep learning.
\newblock \emph{J. Sci. Comput.}, 92\penalty0 (3), September 2022.
\newblock ISSN 0885-7474.
\newblock \doi{10.1007/s10915-022-01911-x}.
\newblock URL \url{https://doi.org/10.1007/s10915-022-01911-x}.

\end{thebibliography}

\newpage
\appendix

\section{Proofs}\label{app:proof}

We provide the proofs of the three main theoretical statements. 
First, we prove that the stochastic version of \ours{} gives an unbiased
estimator of the parameter gradient over a checkpointed segment. The key point
is that each sketched local Jacobian is evaluated at an exact checkpointed
activation, so the masks are independent across blocks after conditioning on
the forward pass. Second, we derive the optimal Bernoulli sampling
probabilities used by Opt-$\frac{k}{n}$ by minimizing the mean-squared
sketching error under a fixed expected budget by deriving the KKT condition of the corresponding optimization problem. Finally, we demonstrate that Top-$\frac{k}{n}$ offers an approximation advantage over Opt-$\frac{k}{n}$.

\subsection{Proof of Proposition~\ref{prop:unbiased-ac-ours}}
\begin{restatedproposition}[Unbiased gradient in the separable setting]
Let $
    \mathcal{L}\circ \Phi_L \circ \cdots \circ \Phi_1
$ be a model and assume that each block $\Phi_l$ admits a separable decomposition as in
Eq.~\eqref{eq:sep-dec}. For each block, let $\omega_l$ be a random mask such
that $\mathbb{E}[\omega_l]=\mathbf{1}$, with masks sampled independently across
blocks. Let $\{x_l^{\mathrm{AC}}\}_{l\leq L}$ be the checkpointed activations obtained from the exact forward pass, as in
Eq.~\eqref{eq:recomputeforward}. Then, the sketched AC backward pass gives an
unbiased  estimator
\[
\mathbb{E}_{\omega}
\left[
\left(
\widehat{\nabla}_{\theta_1}^{\,\omega,\mathrm{AC}}\mathcal{L},
\ldots,
\widehat{\nabla}_{\theta_L}^{\,\omega,\mathrm{AC}}\mathcal{L}
\right)
\right]
=
\nabla_{(\theta_1,\ldots,\theta_L)}
\mathcal{L}\,,
\]
where we have for $1\leq l\leq L$,
\[
\widehat{\nabla}_{\theta_l}^{\,\omega,\mathrm{AC}}\mathcal{L}
:=
\left(
J_{\theta_l}\Phi_l^{\omega_l}(x_l^{\mathrm{AC}},\theta_l)
\right)^{\!\top}
\left(
J_x\Phi_{l+1}^{\omega_{l+1}}(x_{l+1}^{\mathrm{AC}},\theta_{l+1})
\right)^{\!\top}
\cdots
\left(
J_x\Phi_L^{\omega_L}(x_L^{\mathrm{AC}},\theta_L)
\right)^{\!\top}
\nabla_{x_{L+1}}\mathcal{L}.
\]
\end{restatedproposition}
\begin{proof}
For each block $l$, the sketched operator is
\[
\Phi_l^{\omega_l}(x,\theta_l)
=
\sum_i \omega_l^i \varphi_l^i(x,\theta_l^i),
\]
with $\mathbb{E}[\omega_l]=\mathbf{1}$. Hence, for fixed input
$x_l^{\mathrm{AC}}$, both the parameter and input Jacobians are unbiased:
\[
\mathbb{E}_{\omega_l}
\left[
J_{\theta_l}\Phi_l^{\omega_l}(x_l^{\mathrm{AC}},\theta_l)
\right]
=
J_{\theta_l}\Phi_l(x_l^{\mathrm{AC}},\theta_l),
\qquad
\mathbb{E}_{\omega_l}
\left[
J_x\Phi_l^{\omega_l}(x_l^{\mathrm{AC}},\theta_l)
\right]
=
J_x\Phi_l(x_l^{\mathrm{AC}},\theta_l).
\]
The key point is that the activations $x_l^{\mathrm{AC}}$ are obtained from the
exact forward pass and therefore do not depend on the masks $\omega_l$. Since
the masks are independent across blocks, the expectation of the product of the
sketched Jacobians factorizes. Thus, for every $l$,
\[
\begin{aligned}
\mathbb{E}_{\omega}
\left[
\widehat{\nabla}_{\theta_l}^{\,\omega,\mathrm{AC}}\mathcal{L}
\right]
&=
\mathbb{E}_{\omega}
\left[
\left(
J_{\theta_l}\Phi_l^{\omega_l}
\right)^{\!\top}
\left(
J_x\Phi_{l+1}^{\omega_{l+1}}
\right)^{\!\top}
\cdots
\left(
J_x\Phi_L^{\omega_L}
\right)^{\!\top}
\nabla_{x_{L+1}}\mathcal{L}
\right] \\
&=
\left(
J_{\theta_l}\Phi_l
\right)^{\!\top}
\left(
J_x\Phi_{l+1}
\right)^{\!\top}
\cdots
\left(
J_x\Phi_L
\right)^{\!\top}
\nabla_{x_{L+1}}\mathcal{L}.
\end{aligned}
\]
All Jacobians above are evaluated at the exact checkpointed activations
$x_l^{\mathrm{AC}}$ and parameters $\theta_l$. The final expression is
exactly the chain-rule expression for the true gradient
$\nabla_{\theta_l}\mathcal{L}$. Therefore,
\[
\mathbb{E}_{\omega}
\left[
\widehat{\nabla}_{\theta_l}^{\,\omega,\mathrm{AC}}\mathcal{L}
\right]
=
\nabla_{\theta_l}\mathcal{L},
\qquad l=1,\ldots,L,
\]
which proves the claimed unbiasedness of the full gradient estimator.
\end{proof}

\subsection{Proof of Proposition~\ref{prop:opt-bernoulli}}
\begin{restatedproposition}[Optimal Bernoulli sampling probabilities]
Let $\Phi(x)=\sum_{i=1}^n \varphi^i(x,\theta^i)$ and let
\[
    \Phi^\omega(x)=\sum_{i=1}^n \omega^i \varphi^i(x,\theta^i),
    \qquad
    \omega^i=\frac{z_i}{p_i},
    \qquad
    z_i\sim \mathrm{Bernoulli}(p_i),
\]
where the $z_i$ are independent. For a collection of inputs $\{x^m\}_m$, define $\|\varphi^i\|^2
    :=
    \sum_m \|\varphi^i(x^m,\theta^i)\|^2.$
Then the probabilities $p_i$ minimizing Eq.~\eqref{eq:optimal-sampling-problem} are given by
\[
    p_i^\star
    =
    \min\!\left(
        1,
        \max\!\left(
            \varepsilon,
            \frac{\|\varphi^i\|}{\sqrt{\lambda}}
        \right)
    \right),
\]
where $\lambda>0$ is chosen so that
\[
    \sum_{i=1}^n
    \min\!\left(
        1,
        \max\!\left(
            \varepsilon,
            \frac{\|\varphi^i\|}{\sqrt{\lambda}}
        \right)
    \right)
    = k.
\]
In particular, if the box constraints are inactive, i.e.,$
    \varepsilon
    \le
    \frac{k\|\varphi^i\|}
         {\sum_{j=1}^n \|\varphi^j\|}
    \le 1$
    for all  $i$,
 then
\[
    p_i^\star
    =
    \frac{k\|\varphi^i\|}
         {\sum_{j=1}^n \|\varphi^j\|}.
\]
\end{restatedproposition}
\begin{proof}
Using $\omega^i=z_i/p_i$ with $z_i\sim\mathcal B(p_i)$, we have
\[
\mathbb{E}[\omega^i]=1,
\qquad
\mathrm{Var}(\omega^i)=\frac{1-p_i}{p_i}.
\]
Writing $\Vert\varphi^i\Vert^2=\sum_m\Vert\varphi^i(x^m,\theta^i)\Vert^2$, independence gives
\[
\sum_m\mathbb{E}_\omega\|\Phi(x^m)-\Phi^\omega(x^m)\|^2
=
\sum_{i=1}^n \frac{1-p_i}{p_i}\,\|\varphi^i\|^2.
\]
Hence the optimization problem is
\[
\min_{\{p_i\}_i}\sum_{i=1}^n \frac{\|\varphi^i\|^2}{p_i}
\qquad\text{s.t.}\qquad
\sum_{i=1}^n p_i = k,\quad \varepsilon \le p_i \le 1.
\]
The Lagrangian is
\[
\mathcal L(p,\lambda)=\sum_{i=1}^n \frac{\|\varphi^i\|^2}{p_i}
+\lambda\Bigl(\sum_{i=1}^n p_i-k\Bigr).
\]
Ignoring the box constraints for a moment, the stationarity condition gives
\[
-\frac{\|\varphi^i\|^2}{p_i^2}+\lambda=0,
\qquad\text{hence}\qquad
p_i=\frac{\|\varphi^i\|}{\sqrt{\lambda}}.
\]

Therefore, with the constraints $\varepsilon\le p_i\le 1$, the KKT conditions yield
\[
p_i^\star
=
\min\!\left(1,\max\!\left(\varepsilon,\frac{\|\varphi^i\|}{\sqrt{\lambda}}\right)\right),
\]
where $\lambda>0$ is chosen so that
\[
\sum_{i=1}^n
\min\!\left(1,\max\!\left(\varepsilon,\frac{\|\varphi^i\|}{\sqrt{\lambda}}\right)\right)
= k.
\]
In the unconstrained regime, namely when
\[
\varepsilon \le \frac{k\,\|\varphi^i\|}{\sum_{j=1}^n \|\varphi^j\|} \le 1
\qquad\text{for all }i,
\]
this reduces to
\[
p_i^\star
=
\frac{k\,\|\varphi^i\|}{\sum_{j=1}^n \|\varphi^j\|}.
\]
\end{proof}

\subsection{Proof of Proposition~\ref{prop:top-vs-opt}}
\begin{restatedproposition}[Top-$k$ versus optimal Bernoulli sampling]
Let $\omega^{\mathrm{Top}}$ and $\omega^{\mathrm{Opt}}$ be the masks of Top-$\frac kn$ and
Opt-$\frac kn$. Then
\[
Q(\omega^{\mathrm{Top}})
\;\le\;
\bigl(1+(n-k-1)\mu\bigr)\!\!\sum_{i\notin \mathrm{Top}_k}\!\!\|\varphi^i\|^2
\;\le\;
\bigl(1+(n-k-1)\mu\bigr)\,Q(\omega^{\mathrm{Opt}}).
\]
In particular $Q(\omega^{\mathrm{Top}})\le(n-k)\,Q(\omega^{\mathrm{Opt}})$ since $\mu\le1$,
and $Q(\omega^{\mathrm{Top}})\le Q(\omega^{\mathrm{Opt}})$ whenever $\mu=0$. Conversely, no reverse bound holds:
for $n-k\ge2$ there is no constant $C(n,k)$ such that
$Q(\omega^{\mathrm{Opt}})\le C(n,k)\,Q(\omega^{\mathrm{Top}})$.
\end{restatedproposition}

\begin{proof}
Since $\omega^{\mathrm{Top}}$ is deterministic and discards exactly the components outside
$\mathrm{Top}_k$, expanding the squared norm gives
\[
Q(\omega^{\mathrm{Top}})
=\Bigl\|\sum_{i\notin \mathrm{Top}_k}\varphi^i\Bigr\|^2
=\!\!\sum_{i\notin \mathrm{Top}_k}\!\!\|\varphi^i\|^2
+2\!\!\!\sum_{\substack{i<j\\ i,j\notin \mathrm{Top}_k}}\!\!\!\langle\varphi^i,\varphi^j\rangle
\;\le\;
\!\!\sum_{i\notin \mathrm{Top}_k}\!\!\|\varphi^i\|^2
+2\mu\!\!\!\sum_{\substack{i<j\\ i,j\notin \mathrm{Top}_k}}\!\!\!\|\varphi^i\|\,\|\varphi^j\| ,
\]
by definition of $\mu$. Young's inequality $2\|\varphi^i\|\|\varphi^j\|\le\|\varphi^i\|^2
+\|\varphi^j\|^2$, summed over the unordered pairs outside $\mathrm{Top}_k$, makes each
term appear $n-k-1$ times and yields the first inequality.

For the second, recall from the proof of Prop.~\ref{prop:opt-bernoulli} that
$Q(\omega)=\sum_i\frac{1-p_i}{p_i}\|\varphi^i\|^2$. Any feasible $p$ satisfies $p_i\le1$,
so $\frac{1-p_i}{p_i}\ge 1-p_i$ and
\[
Q(\omega)\;\ge\;\sum_{i=1}^n\|\varphi^i\|^2-\sum_{i=1}^n p_i\|\varphi^i\|^2 .
\]
The linear functional $p\mapsto\sum_i p_i\|\varphi^i\|^2$ over
$\{0\le p_i\le1,\ \sum_i p_i=k\}$ is maximized at the vertex
$p=\mathbf 1_{\mathrm{Top}_k}$, hence
$Q(\omega)\ge\sum_{i\notin \mathrm{Top}_k}\|\varphi^i\|^2$ for every feasible $p$, and in
particular for $\omega^{\mathrm{Opt}}$. Finally $\mu\le1$ by Cauchy--Schwarz, and $\mu=0$
under non-positive inner products.

For the last claim, let the $n-k$ components outside $\mathrm{Top}_k$ be unit vectors
summing to zero and the $k$ others have norm $1+\epsilon$, so that Top-$\frac kn$ indeed
retains the latter. Then $Q(\omega^{\mathrm{Top}})=0$, while $\sum_ip_i=k<n$ forces
$p_i<1$ for some $i$ and hence $Q(\omega^{\mathrm{Opt}})>0$.
\end{proof}

\section{AC-FLOP derivations}
\label{sec:app-flops}
Let $n_{\mathrm{FFN}}$ denote the FFN hidden size and
$n_{\mathrm{Attention}}$ the number of attention heads. From Sec.~\ref{sec:main-metod-kweak},
the FLOPs for the FFN and attention layers without AC are
\begin{equation}
18 n_{\mathrm{FFN}}D + 24D^2 .
\end{equation}

Applying \ours{} to both FFN and attention layers at token-level granularity
gives
\begin{equation}
\left(6n_{\mathrm{FFN}}D + 18
\frac{k_{\mathrm{FFN}}}{n_{\mathrm{FFN}}}n_{\mathrm{FFN}}D\right)
+
\left(16D^2 + 12
\frac{k_{\mathrm{Attention}}}{n_{\mathrm{Attention}}}D^2\right).
\end{equation}

Let
$p=\frac{k_{\mathrm{Attention}}}{n_{\mathrm{Attention}}}
=\frac{k_{\mathrm{FFN}}}{n_{\mathrm{FFN}}}$.
The FLOP ratio between \ours{} and the vanilla backward pass is then
\begin{equation}
\frac{
6n_{\mathrm{FFN}}D + 18p n_{\mathrm{FFN}}D
+ 16D^2 + 12pD^2
}{
18n_{\mathrm{FFN}}D + 24D^2
}.
\end{equation}
For $n_{\mathrm{FFN}}=\frac{8}{3}D$, the formula becomes:
\begin{equation}
\frac{
6\times \frac{8}{3}D^2
+ 18p \times \frac{8}{3}D^2
+ 16D^2
+ 12pD^2
}{
18\times \frac{8}{3}D^2
+ 24D^2
}.
\end{equation}

Simplifying,
\begin{equation}
\frac{
16D^2 + 48pD^2 + 16D^2 + 12pD^2
}{
48D^2 + 24D^2
}
=
\frac{
32D^2 + 60pD^2
}{
72D^2
}.
\end{equation}

Canceling $D^2$ gives
\begin{equation}
\frac{32 + 60p}{72}
=
\frac{4}{9} + \frac{5}{6}p.
\end{equation}

Substituting $p=\frac{1}{3}$ and $p=\frac{2}{3}$ gives FLOP ratios of approximately $0.72$ and $1$, respectively.

\section{1B-Parameter Experiment}
\label{app:1b-experiment}
To assess whether KEAP remains effective at a larger model scale, we train a Llama-3-style model with approximately one billion parameters. The model has 953M parameters and consists of 16 Transformer layers, a hidden dimension of 2,048, and 32 attention heads. It follows the same training setup described in Sec.~\ref{sec:training-setup} and is evaluated on the same held-out FineWeb set used in the experiments of Section~\ref{sec:main-results}. We reuse the learning rate tuned for the 488M-parameter dense baseline
for both the dense and KEAP 953M runs, without additional tuning. For KEAP, we apply token-level Top-$\frac{2}{3}$ selection exclusively to the FFN layers. The results are reported in Table~\ref{tab:1b-results}.

\begin{table}[t]
\centering
\small
\captionsetup{skip=2pt}
\begin{tabular}{lcc}
\toprule
Method & CE & PPL \\
\midrule
Dense & 2.288 & 9.858 \\
KEAP FFN Top-$\frac{2}{3}$ & 2.298 & 9.959 \\
\bottomrule
\end{tabular}
\caption{\textbf{KEAP scaling performance}
We report the cross-entropy loss and perplexity of a vanilla Llama-953M model versus KEAP Top-$\frac{2}{3}$ with token-level granularity applied to the FFN}
\label{tab:1b-results}
\end{table}

KEAP increases perplexity from $9.858$ to $9.959$, corresponding to a relative degradation of only $1.02\%$. This is smaller than the $1.66\%$ relative degradation observed for the corresponding FFN Top-$\frac{2}{3}$ token-level setting on the 488M-parameter model. 

On four H100 GPUs, with a sequence length of 2,048 and a local batch size of 16 per GPU, the 953M-parameter model uses $76.5$ GiB per GPU without activation checkpointing and $31.0$ GiB with Full AC. In this setting, KEAP requires only an additional $11$ KiB per GPU for batch-level selection and $0.34$ GiB per GPU for token-level selection when the forward-pass selection masks are stored using one bit per decision.

\section{Results with Block-granularity}
\label{sec:main-block}
The experiments reported in Table~\ref{tab:batch-level} use a single mask shared across all samples in a batch, allowing the retained FFN computation to be executed using standard dense GEMMs. For instance, on the 953M-parameter model described in Appendix ~\ref{app:1b-experiment}, trained on four H100 GPUs with a sequence length of 2,048 and a local batch size of 16 per GPU, batch-level FFN KEAP reduces training-step time by $26\%$ with Top-$\frac{1}{3}$ and by $18\%$ with Top-$\frac{2}{3}$ relative to Full AC. However, token-level selection generally achieves better accuracy, as shown in Tables~\ref{tab:token-level} and~\ref{tab:batch-level}, because each token independently selects its most important FFN channels. However, this accuracy comes at the cost of less regular computation, since different tokens may retain different channels.

We explore two complementary ways of navigating this accuracy--efficiency trade-off. First, a mask can be shared across groups of tokens, gradually moving from fine-grained token-level selection toward more structured batch-level selection and producing larger, more regular matrix operations. Second, FFN channels can be partitioned into blocks, with entire blocks selected jointly. Tokens that retain the same blocks can then be packed and processed using grouped GEMMs. Larger channel blocks generally improve computational efficiency, whereas smaller or infrequently selected blocks remain more sensitive to dispatch, memory-movement, and kernel-launch overhead. We therefore study token-group and channel-group sizes jointly, spanning a continuum between fine-grained token-level KEAP and structured batch-level KEAP while keeping the retained fraction $k/n$ fixed. To our knowledge, this joint grouping strategy has not previously been studied for approximate activation-checkpoint rematerialization. These block-granularity accuracy ablations were conducted on NVIDIA V100 GPUs using FP16 mixed precision.

\begin{figure}[t]
\centering
\includegraphics[width=0.49\linewidth]{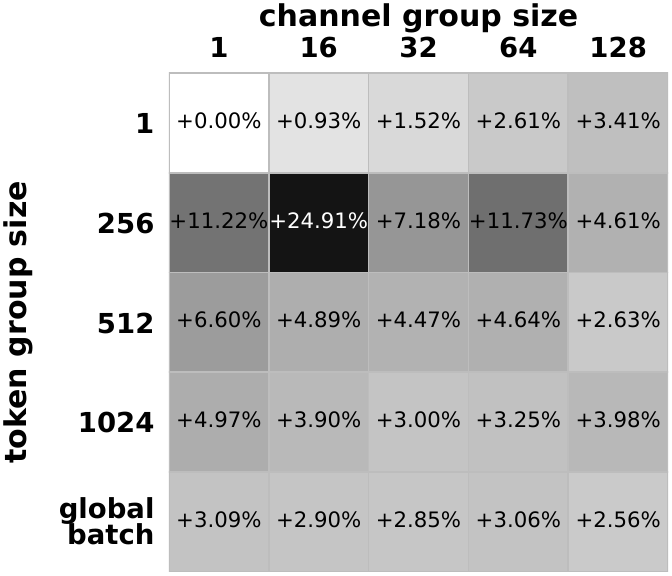}
\includegraphics[width=0.49\linewidth]{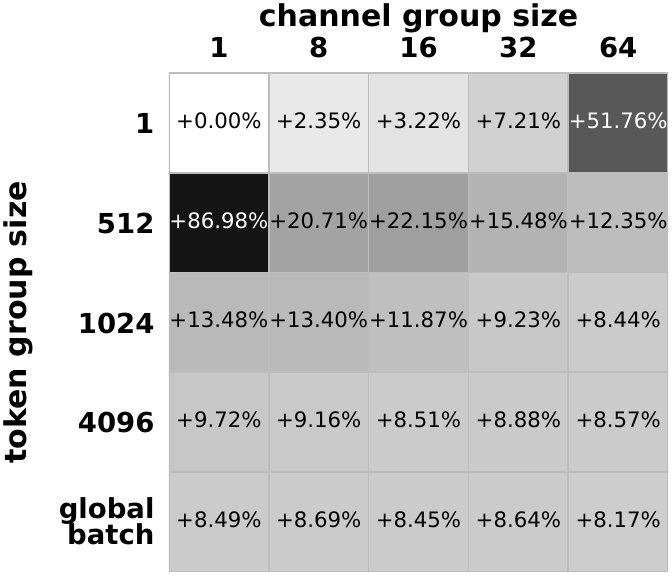}
\caption{\textbf{Effect of token and channel grouping in KEAP FFN.} We train a Llama-134M model while varying the token-group size and FFN channel-group size. Each cell reports the relative perplexity degradation with respect to per-token, per-channel KEAP at the same sparsity fraction. Left: Top-$\frac{2}{3}$. Right: Top-$\frac{1}{3}$.}
\label{fig:keap_blockwise_ablation}
\end{figure}

Figure~\ref{fig:keap_blockwise_ablation} shows the resulting accuracy trade-off. For Top-$\frac{2}{3}$, channel-group sizes of 16 and 32 under token-level selection increase relative perplexity by only $0.93\%$ and $1.52\%$, respectively. As the token-group size increases, performance approaches that of global-batch selection. For example, a token-group size of 512 combined with a channel-group size of 128 yields a $2.63\%$ increase in relative perplexity, compared with $2.56\%$ for global-batch selection at the same channel-group size. Top-$\frac{1}{3}$ is more sensitive to the grouping configuration, although larger token groups similarly approach the global-batch result. The corresponding speedups depend strongly on the target hardware and the implementation of data movement and grouped kernels, making their systematic characterization an important direction for future work.

\section{Evaluation Loss}

We present the results from Tab.~\ref{tab:remat-results} as their corresponding cross entropy values in Tab.~\ref{tab:ce-results}
\begin{table*}[t]
\centering
\small
\setlength{\tabcolsep}{3pt}

\begin{subtable}[t]{0.45\textwidth}
\centering
\begin{tabular}{lccc}
\toprule
Method &  $\frac kn$ & AC-FLOPs & CE \\
\midrule
\multicolumn{4}{l}{\footnotesize AC applied on FFN layers} \\
\midrule
Llama-488M & 1 & 1.33 & 2.487 \\
Llama-419M & 1 & 1 & 2.521 \\
\midrule
Rand & 0.66 & 1 & 2.553 \\
     & 0.33 & 0.66 & 2.688 \\
\midrule
Opt  & 0.66 & 1 & 2.532 \\
     & 0.33 & 0.66 & 2.614 \\
\midrule
Top  & 0.66 & 1 & \textbf{2.504} \\
     & 0.33 & 0.66 & 2.523 \\
\midrule
\multicolumn{4}{l}{\footnotesize AC applied on Attention layers} \\
\midrule
Llama-488M & 1 & 1.33 & 2.487 \\
Llama-468M & 1 & 1 & \textbf{2.499} \\
\midrule
Rand & 0.66 & 1 & 2.530 \\
     & 0.33 & 0.83 & 2.598 \\
\midrule
Opt  & 0.66 & 1 & 2.507 \\
     & 0.33 & 0.83 & 2.545 \\
\midrule
Top  & 0.66 & 1 & \textbf{2.499} \\
     & 0.33 & 0.83 & 2.522 \\
\midrule
\multicolumn{4}{l}{\footnotesize AC applied on Attention+FFN layers} \\
\midrule
Llama-488M & 1 & 1.33 & 2.487 \\
Llama-400M & 1 & 1 & 2.531 \\
\midrule
Rand & 0.66 & 1 & 2.601 \\
     & 0.33 & 0.72 & 2.885 \\
\midrule
Opt  & 0.66 & 1 & 2.550 \\
     & 0.33 & 0.72 & 2.696 \\
\midrule
Top  & 0.66 & 1 & \textbf{2.512} \\
     & 0.33 & 0.72 & 2.557 \\
\bottomrule
\end{tabular}
\caption{Token-level granularity}
\label{tab:token-level-ce}
\end{subtable}
\hfill
\begin{subtable}[t]{0.45\textwidth}
\centering
\begin{tabular}{lccc}
\toprule
Method & $\frac kn$ & AC-FLOPs & CE \\
\midrule
\multicolumn{4}{l}{\footnotesize AC applied on FFN layers} \\
\midrule
Llama-488M & 1 & 1.33 & 2.487 \\
Llama-419M & 1 & 1 & \textbf{2.521} \\
\midrule
Rand & 0.66 & 1 & 2.570 \\
     & 0.33 & 0.66 & 2.744 \\
\midrule
Opt  & 0.66 & 1 & 2.557 \\
     & 0.33 & 0.66 & 2.660 \\
\midrule
Top  & 0.66 & 1 & 2.534 \\
     & 0.33 & 0.66 & 2.595 \\
\midrule
\multicolumn{4}{l}{\footnotesize AC applied on Attention layers} \\
\midrule
Llama-488M & 1 & 1.33 & 2.487 \\
Llama-456M & 1 & 1 & \textbf{2.503} \\
\midrule
Rand & 0.66 & 1 & 2.562 \\
     & 0.33 & 0.66 & 2.672 \\
\midrule
Opt  & 0.66 & 1 & 2.549 \\
     & 0.33 & 0.66 & 2.622 \\
\midrule
Top  & 0.66 & 1 & 2.514 \\
     & 0.33 & 0.66 & 2.529 \\
\midrule
\multicolumn{4}{l}{\footnotesize AC applied on Attention+FFN layers} \\
\midrule
Llama-488M & 1 & 1.33 & 2.487 \\
Llama-387M & 1 & 1 & \textbf{2.533} \\
\midrule
Rand & 0.66 & 1 & 2.657 \\
     & 0.33 & 0.66 & 3.007 \\
\midrule
Opt  & 0.66 & 1 & 2.624 \\
     & 0.33 & 0.66 & 2.837 \\
\midrule
Top  & 0.66 & 1 & 2.553 \\
     & 0.33 & 0.66 & 2.649 \\
\bottomrule
\end{tabular}
\caption{Batch-level granularity}
\label{tab:batch-level-ce}
\end{subtable}
\caption{\textbf{\ours{} performance}.
We report the cross entropy loss of a Llama-488M model under different AC strategies, using per-token (\textbf{a}) and per-batch (\textbf{b}) granularity  during re-computation as described in Sec.~\ref{sec:implem}. AC-FLOPs denotes the ratio between the FLOP cost of each AC method and the cost of standard backpropagation. (details can be found in  App.~\ref{sec:app-flops})}
\label{tab:ce-results}
\end{table*}

\label{app:eval}

\section{Assets and licenses}
We use FineWeb \citep{penedo2024fineweb}, available at
\url{https://huggingface.co/datasets/HuggingFaceFW/fineweb}, under the
ODC-By 1.0 license. We use TorchTitan, available at
\url{https://github.com/pytorch/torchtitan}, under the BSD-3-Clause license.
We use the Llama-2 7B tokenizer, available through the Llama-2 model release,
under the Meta Llama 2 Community License Agreement:
\url{https://ai.meta.com/llama/license/}. All models are trained from
scratch; we do not use or redistribute pretrained model weights.

\section{Compute resources}
Each run reported in the main results of Sec.~\ref{sec:main-results} required less than 30 H100-equivalent GPU-hours with an efficient implementation. The total compute required for the reported main-result experiments was below 1000 H100-equivalent GPU-hours.

\section{Limitations}
Our main evaluation focuses on Llama-style models up to 488M parameters, supplemented by a 953M-parameter scaling experiment, and results may differ at substantially larger scales, on other architectures, or on other data distributions. Most results are based on single training runs due to computational cost, so we do not report run-to-run variance. For attention rematerialization with a token level selector we mention that key-value computations constrain the achievable savings. 

\section{Broader Impacts} 
This work is a training-efficiency method whose primary broader impact is to reduce the compute and energy overhead of large-model training. More efficient training may also lower the barrier to training larger models, including models used in harmful or poorly governed applications. We do not identify a direct path to negative societal applications beyond this generic dual-use effect of improving neural-network training efficiency.

\newpage

\section*{NeurIPS Paper Checklist}

\begin{enumerate}

\item {\bf Claims}
    \item[] Question: Do the main claims made in the abstract and introduction accurately reflect the paper's contributions and scope?
    \item[] Answer: \answerYes{} 
    \item[] Justification: The main claims stated in the abstract and introduction are supported by both the theoretical analysis and experiments. The main empirical results are based on a tuned 488M-parameter Llama-style model trained with a compute-optimal token budget. The claims are limited to the model scales studied.

\item {\bf Limitations}
    \item[] Question: Does the paper discuss the limitations of the work performed by the authors?
    \item[] Answer: \answerYes{} 
    \item[] Justification: The paper includes a limitations section discussing the scope of the evaluation, which is limited to Llama-style models up to 1B parameters trained on FineWeb. It also discusses the use of single training runs and the constrained savings for token level \ours{} for attention due to key-value computations under token-level selection.

\item {\bf Theory assumptions and proofs}
    \item[] Question: For each theoretical result, does the paper provide the full set of assumptions and a complete (and correct) proof?
    \item[] Answer:\answerYes{} 
    \item[] Justification: The paper states the conditions under which the theoretical results hold and complete proofs are provided in the appendix.

    \item {\bf Experimental result reproducibility}
    \item[] Question: Does the paper fully disclose all the information needed to reproduce the main experimental results of the paper to the extent that it affects the main claims and/or conclusions of the paper (regardless of whether the code and data are provided or not)?
    \item[] Answer: \answerYes{} 
    \item[] Justification: The paper provides sufficient details on model architecture, training setup, hyperparameters, and evaluation protocol to reproduce the main results.

\item {\bf Open access to data and code}
    \item[] Question: Does the paper provide open access to the data and code, with sufficient instructions to faithfully reproduce the main experimental results, as described in supplemental material?
    \item[] Answer: \answerYes{} 
    \item[] Justification: The data used in the experiments are publicly available, and the paper describes the data source and training setup. The implementation code is also shared.

\item {\bf Experimental setting/details}
    \item[] Question: Does the paper specify all the training and test details (e.g., data splits, hyperparameters, how they were chosen, type of optimizer) necessary to understand the results?
    \item[] Answer: \answerYes{} 
    \item[] Justification: The experimental section specifies datasets, model configurations, training procedure, and optimization details needed to understand the results. Furthermore, these details follow the protocol of another peer reviewed paper. 

\item {\bf Experiment statistical significance}
    \item[] Question: Does the paper report error bars suitably and correctly defined or other appropriate information about the statistical significance of the experiments?
    \item[] Answer: \answerNo{} 
    \item[] Justification: Results are based on single training runs due to computational cost; error bars or variance estimates are not reported.

\item {\bf Experiments compute resources}
    \item[] Question: For each experiment, does the paper provide sufficient information on the computer resources (type of compute workers, memory, time of execution) needed to reproduce the experiments?
    \item[] Answer: \answerYes{} 
    \item[] Justification: The paper reports the GPU types, number of GPUs, numerical precision, model sizes, token budgets, memory and FLOP analyses, and an upper bound on GPU-hours for the reported main-result experiments.
    
\item {\bf Code of ethics}
    \item[] Question: Does the research conducted in the paper conform, in every respect, with the NeurIPS Code of Ethics \url{https://neurips.cc/public/EthicsGuidelines}?
    \item[] Answer: \answerYes{} 
    \item[] Justification: The work conforms to the NeurIPS Code of Ethics. It does not involve human subjects, crowdsourcing, surveillance data, security vulnerabilities, or the release of high-risk models. The experiments use an existing public language-model training dataset and standard training software, and the paper reports relevant compute and reproducibility details.

\item {\bf Broader impacts}
    \item[] Question: Does the paper discuss both potential positive societal impacts and negative societal impacts of the work performed?
    \item[] Answer: \answerYes{} 
    \item[] Justification: A dedicated section in the appendix discusses the positive impact of reducing compute and energy overhead for large-model training, as well as the generic dual-use concern that more efficient training may lower the barrier to training larger models, including models used in harmful or poorly governed applications.
    
\item {\bf Safeguards}
    \item[] Question: Does the paper describe safeguards that have been put in place for responsible release of data or models that have a high risk for misuse (e.g., pre-trained language models, image generators, or scraped datasets)?
    \item[] Answer: \answerNA{} 
    \item[] Justification: The paper does not release a pretrained model, image generator, scraped dataset, or other high-risk asset requiring release safeguards. The contribution is a training-efficiency method.

\item {\bf Licenses for existing assets}
    \item[] Question: Are the creators or original owners of assets (e.g., code, data, models), used in the paper, properly credited and are the license and terms of use explicitly mentioned and properly respected?
    \item[] Answer: \answerYes{} 
    \item[] Justification: In a dedicated section in the appendix the paper cites the existing datasets, software, and model/tokenizer assets used in the experiments and states that they are used under their respective licenses and terms of use.

\item {\bf New assets}
    \item[] Question: Are new assets introduced in the paper well documented and is the documentation provided alongside the assets?
    \item[] Answer: \answerYes{} 
    \item[] Justification: Source code is released.

\item {\bf Crowdsourcing and research with human subjects}
    \item[] Question: For crowdsourcing experiments and research with human subjects, does the paper include the full text of instructions given to participants and screenshots, if applicable, as well as details about compensation (if any)? 
    \item[] Answer: \answerNA{} 
    \item[] Justification: The work does not involve crowdsourcing, user studies, or human-subject experiments.

\item {\bf Institutional review board (IRB) approvals or equivalent for research with human subjects}
    \item[] Question: Does the paper describe potential risks incurred by study participants, whether such risks were disclosed to the subjects, and whether Institutional Review Board (IRB) approvals (or an equivalent approval/review based on the requirements of your country or institution) were obtained?
    \item[] Answer: \answerNA{} 
    \item[] Justification: The work does not involve human subjects, so IRB approval or equivalent review is not applicable.

\item {\bf Declaration of LLM usage}
    \item[] Question: Does the paper describe the usage of LLMs if it is an important, original, or non-standard component of the core methods in this research? Note that if the LLM is used only for writing, editing, or formatting purposes and does \emph{not} impact the core methodology, scientific rigor, or originality of the research, declaration is not required.
    \item[] Answer: \answerNA{} 
    \item[] Justification: LLMs are not used as part of the core method, experiments, or scientific contribution. Any use is limited to writing or editing assistance and does not affect the methodology or results.

\end{enumerate}

\end{document}